\documentclass{article}

\usepackage[main,final]{neurips_2024}

\usepackage[utf8]{inputenc} 
\usepackage[T1]{fontenc}    
\usepackage{hyperref}       
\usepackage{url}            
\usepackage{booktabs}       
\usepackage{amsfonts}       
\usepackage{amsmath}        
\usepackage{nicefrac}       
\usepackage{microtype}      
\usepackage[dvipsnames,table]{xcolor}         
\usepackage{wrapfig}
\usepackage{xspace}

\usepackage{graphicx}
\usepackage{multirow}
\usepackage{subcaption}

\definecolor{groupgray}{gray}{0.92}
\definecolor{avgblue}{RGB}{232,236,240} 
\definecolor{mygreen}{RGB}{60,110,80}   

\definecolor{myred}{RGB}{140,75,75}   

\usepackage{needspace}

\usepackage[normalem]{ulem}

\newcommand{\molfull}{\textbf{\textsc{Mixture of Layers}}}
\newcommand{\molshort}{\textbf{\textsc{MoL}}}

\title{\textsc{Mixture of Layers}:\\Dynamic Layer Routing for Visual Reasoning}
\author{%
 Jeonghwan Kim$^{1}$\thanks{~Equal contribution.}, Sofia Stoica$^{1*}$, Jiwan Chung$^{2}$, Ansel Blume$^{1}$, \\
 \bf Hyeonjeong Ha$^{1}$, Zhenhailong Wang$^{1}$, Xin Luna Dong$^{3}$, Heng Ji$^{1}$\\
$^{1}$University of Illinois Urbana-Champaign, $^{2}$Yonsei University, $^{3}$Meta Reality Labs \\
\texttt{\{jk100, sstoica2, hengji\}@illinois.edu}
}

\begin{document}

\maketitle

\begin{abstract}
  Pre-trained vision encoders contain layer-wise visual representations that differ in spatial granularity, semantic abstraction, and sensitivity to local details. However, most Multimodal Large Language Models (MLLMs) rely on only the final or penultimate vision encoder representations or fixed aggregation rules, making visual abstraction largely query-agnostic and limiting access to fine-grained cues such as small objects, spatial details, text, and subtle visual attributes.
  In this work, we propose \molfull~(\molshort), an instruction-conditioned layer routing approach at visual patch level that dynamically aggregates query-relevant latent representations from intermediate vision encoder layers. Given a text query, \molshort~predicts routing probabilities over vision encoder layers and performs a top-$k$ sparse aggregation over selected hidden states at either the image level, patch level, or through a hybrid routing mechanism. In doing so, \molshort~enables query-adaptive access to layer-specific visual features for \textit{fine-grained} visual reasoning.
  Our experiments across 7 fine-grained visual reasoning tasks demonstrate substantial performance improvements, especially across fine-grained visual grounding and understanding 
  tasks such as +18.9\% improvement on \textit{V*} in overall accuracy, +4.5\% on HRBench4K, and +16.3\% on CharXiv compared to the baseline MLLMs, \textit{without} resorting to multi-resolution inputs, simple interleaving of multiple vision encoders, or increasing the number of patch tokens. We study vision encoders' receptive field scales across different layers and their sampling behaviors to provide an in-depth analysis of why layer-wise sampling is helpful, demonstrating that conditional visual representations are a key step towards better visual perception and reasoning in MLLMs. Our project page is available at \url{https://wjdghks950.github.io/mol.github.io/}.
\end{abstract}

\section{Introduction}
\label{sec:introduction}


Multimodal Large Language Models (MLLMs) process text and images as input, with designs that typically contain three components: one or more pre-trained vision encoders \citep{radford2021clip, zhai2023sigmoid, oquabdinov2, he2020momentum, heinrich2025radiov2}, a vision-language connector \citep{liu2023llava, koh2023grounding, shi2025eagle, dai2023instructblip, li2023blip, li2025tokenpacker, cha2024honeybee} that maps the visual features extracted from the vision encoder to the pre-trained Large Language Model (LLM) space \citep{li2023textbooks, touvron2023llama, grattafiori2024llama}, and the backbone LLM that takes as input the visual patches and textual tokens to perform complex visual understanding and reasoning. Despite their strong image-text understanding capabilities, many existing MLLMs typically assume that the \textit{later layer} (e.g., final or penultimate) vision encoder representations provide a sufficiently universal abstraction for all downstream tasks as they are strongly aligned with semantic and language-level supervision during training. 


Such assumptions, however, become especially limiting for \textit{fine-grained multimodal perception and reasoning}, where answers often depend on small objects, localized text, subtle attributes, counting, or spatial relations that may not be preserved in a single final-layer representation. We argue that such fixed set of later layers provide limited, query-agnostic access to such granular visual information, even though prior analyses show that spatial and semantic information is distributed differently across vision-encoder layers \citep{gandelsmaninterpreting, ghiasi2022vision, chen2024invite, jiang2024from, bolya2025perceptionencoderbestvisual}. Even query-conditioned visual extraction approaches such as Q-Former-based architectures (e.g., BLIP-2~\citep{li2023blip}, InstructBLIP~\citep{dai2023instructblip}) or Perceiver Resampler-based designs (e.g., Flamingo~\citep{alayrac2022flamingo}) primarily operate on fixed final-layer vision representations rather than dynamically selecting among intermediate-layer features.
Recent analyses show that pretrained vision encoders such as CLIP \citep{radford2021clip} and DINOv2 \citep{oquabdinov2} encode diverse visual features across layers \citep{gandelsmaninterpreting, chen2024invite, jiang2024from}, with the most useful layer varying by downstream task \citep{bolya2025perceptionencoderbestvisual}. 
While recent methods exploit intermediate-layer representations \citep{yao2024dense, cao2024mmfuser, lin2025multi, chen2025multimodal, li2026instruction, meng2024deepstack, lin2026text}, they typically rely on manually selected fixed layers \citep{he2024multi, lin2025multi, cao2024mmfuser, chen2025multimodal}, increase the number of injected visual tokens \citep{yao2024dense, lin2025multi}, or conduct weighted averaging across the layers \citep{lin2026text}. More importantly, these choices are largely \textit{text-instruction agnostic}: the same set of representations are used regardless of visual reasoning task (i.e., prompts). Finally, such static treatment also contrasts with biological vision, where the brain appears to selectively emphasize task-relevant levels of this hierarchy rather than use all representations uniformly \citep{Steel5723} . This suggests that effective visual processing in MLLMs should dynamically select the representations most relevant to the current instruction, rather than simply exposing the LLM to more visual features or fixed end-layer representations.

Inspired by this principle, we propose \molfull~(\molshort), an instruction-guided layer routing mechanism that dynamically selects intermediate-layer visual patch representations from MLLM vision encoders. Rather than relying on a fixed set of visual features, \molshort{} predicts instruction-conditioned routing distributions over vision-encoder layers, enabling different image regions and tasks to access different levels of visual abstraction. Conceptually, the router treats intermediate vision layers as a set of candidate representations and selectively aggregates the most \textit{useful} ones for the current query. We instantiate \molshort{} with three routing variants: layer-level routing (\molshort$_{layer}$), patch-level routing (\molshort$_{patch}$), and hybrid routing (\molshort$_{hybrid}$), which combines the preceding two variants. Specifically, \molshort{}$_{layer}$ performs image-level routing across the layers of the pre-trained vision encoder, while \molshort$_{patch}$ predicts a separate layer distribution for each visual patch, allowing different image regions to draw from different abstraction levels within the vision hierarchy. \molshort$_{hybrid}$ combines the two routing schemes through a weighted product-of-experts~\citep{hinton1999products} and uses their disagreement to gate a reserve layer (\S\ref{sec:preliminary}), balancing local fine-grained evidence with broader semantic context. \molshort{} is particularly effective in fine-grained visual reasoning, showing significant improvements on downstream tasks such as \textit{V*} and CharXiv with gains of 18.9\% and 16.3\%, respectively. In practice, the added cost is small relative to the overall model: the router contributes negligible parameter overhead (<0.02\% over the 13B base model) while leaving token-by-token language decoding unchanged.

 Our work makes three key contributions:
 {\setlength{\itemsep}{2pt}
 \setlength{\parskip}{1pt}
 \setlength{\parsep}{0pt}
 \begin{itemize}
     \item We introduce a novel routing formulation that adaptively selects intermediate vision-encoder layer representations conditioned on the input instruction, supporting both image-level \molshort$_{layer}$ and patch-level \molshort$_{patch}$ decisions.
     \item We develop a hybrid routing mechanism \molshort$_{hybrid}$ with a disagreement-gated reserve layer (\S\ref{sec:preliminary}) that balances local fine-grained evidence and global semantic context while routing across vision encoder layers.
     \item Across seven different fine-grained visual reasoning benchmarks, \molshort{} substantially improves fine-grained visual reasoning without increasing the number of visual tokens passed into the LLM or relying on multiple vision encoders, making \molshort{} complementary to existing MLLM scaling approaches.
 \end{itemize}
 }

\section{Related Work}

\paragraph{Multimodal Large Language Models}
Multimodal Large Language Models (MLLMs) ingest visual patch tokens and textual prompts to generate textual responses \citep{liu2023llava, zhu2024minigpt, li2025llavaonevision, bai2025qwen3, dai2023instructblip, li2023blip, deitke2025molmo}. Most MLLMs rely on a pre-trained vision encoder, a vision-language connector that projects visual features into the LLM input space, and a pre-trained LLM backbone \citep{liu2023llava, zhu2024minigpt, li2025llavaonevision, bai2025qwen3, deitke2025molmo}. 
While architectures such as BLIP-2 \citep{li2023blip}, InstructBLIP \citep{dai2023instructblip}, and Flamingo \citep{alayrac2022flamingo} introduce text-conditioned querying modules such as Q-Former or Perceiver Resampler, they still operate on fixed final or penultimate-layer vision representations. Similarly, visual abstractors and resamplers in models such as mPLUG-OWL \citep{ye2023mplug} and Honeybee \citep{cha2024honeybee} compress or transform already-extracted visual embeddings rather than dynamically selecting intermediate-layer features. 

Consequently, most MLLMs continue to rely on the final or penultimate layers of pre-trained vision encoders \citep{radford2021clip, zhai2023sigmoid, oquabdinov2} under the assumption that these representations sufficiently capture the visual information required for downstream reasoning tasks. 
In contrast, \molshort{} performs instruction-conditioned routing \emph{across} intermediate vision-encoder layers, enabling different image regions and tasks to access different levels of visual abstraction. We leave aside vision encoder-agnostic frameworks such as Emu-3.5 \citep{cui2025emu3} and SOLO \citep{chen2024solo}, since they do not integrate pre-trained vision encoders.

\paragraph{Intermediate Layer Sampling from Pre-trained Vision Encoders}
Prior work evaluating the beneficial effect of intermediate visual features on MLLMs' visual reasoning \citep{cao2024mmfuser, yao2024dense, lin2025multi, chen2025multimodal, li2026instruction} shares a similar motivation to our proposed method: look into the intermediate layers' hidden states to retrieve useful visual information. However, they focus on either rule-based selection or visual feature-only sampling strategies, limiting their applicability to novel, fine-grained downstream tasks. \molshort~overcomes prior static approaches to intermediate representations through instruction-guided sampling at multiple levels of representational granularity (i.e., patch- and layer-level). Concurrent work such as TGIF~\citep{lin2026text} predicts layer weights from the query through an MLP and weighted averages all vision layers. In contrast, MoL\(_{\text{layer}}\) directly measures query-layer representation compatibility and performs sparse top-\(k\) routing, allowing the selected abstraction depth to depend on the visual evidence present in the input rather than only on the semantics of the prompt.

\paragraph{Fine-grained Visual Perception and Reasoning}
Most widely used visual reasoning benchmarks (e.g., MMMU~\citep{yue2024mmmu}, TextVQA~\citep{singh2019textvqa}, POPE~\citep{li2023evaluating}, GQA~\citep{hudson2019gqa}) primarily evaluate coarse-grained understanding and do not stress fine-grained perception in dense, high-resolution scenes critical in real-world tasks such as industrial inspection or document/diagram understanding. Recent benchmarks such as \textit{V*}~\citep{wu2024v}, HRBench~\citep{wang2025divide}, NaturalBench~\citep{li2024naturalbench}, RealWorldQA~\citep{xai-realworldqa}, and CharXiv~\citep{wang2024charxiv} address this gap by requiring models to reason over localized text, small objects, and subtle attributes.

Prior approaches to improving fine-grained reasoning largely focus on increasing the visual token budget via multi-resolution or native-resolution inputs~\citep{li2025llavaonevision, meng2024deepstack, bai2025qwen3}, or concatenating patch embeddings from multiple pre-trained vision encoders~\citep{shi2025eagle, Tong_2024_CVPR, jiang2024from}. By contrast, \molshort~improves fine-grained reasoning by refining the representation of each patch through instruction-conditioned selection of intermediate-layer features, without introducing additional visual tokens or simply concatenating features from multiple vision encoders \citep{shi2025eagle, Tong_2024_CVPR, jiang2024from}. This positions \molshort~as a simple, complementary approach to enhance visual representation quality without scaling input size, allowing it to be integrated with existing multi-patch, high-resolution approaches.
\section{\textsc{Mixture of Layers}: Selective Visual Representations for Fine-Grained Visual Reasoning}
\label{sec:method}

\begin{figure*}[t]
    \centering
    \includegraphics[width=\linewidth]{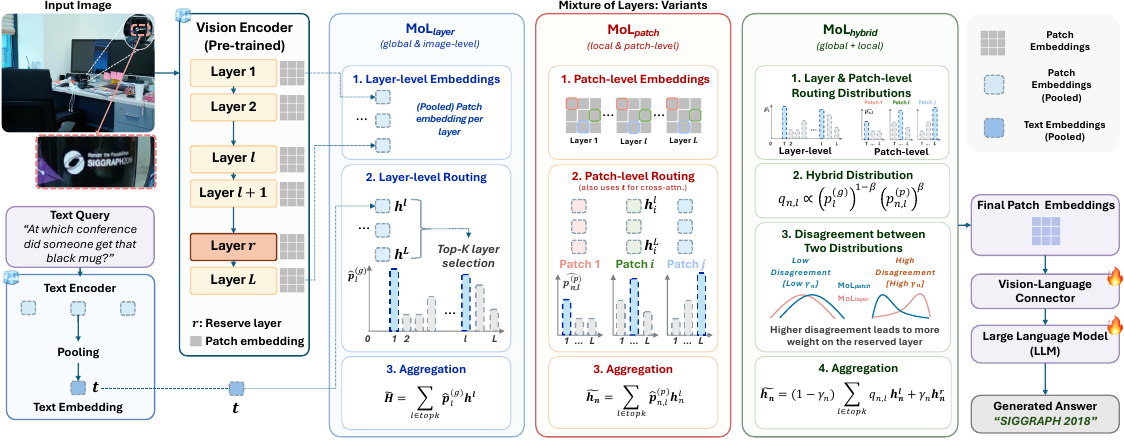}
    \caption{\textbf{Overview of \molfull.} This figure illustrates a fine-grained visual reasoning scenario where the MLLM should ground its reasoning based on visual representations that accurately capture the query-relevant visual features. \molshort~can be divided into three complementary variants: (i) \molshort$_{layer}$ that generates a query-guided ($\bar{t}$) layer-level probability distribution over the input image; (ii) \molshort$_{patch}$, which generates a query-guided distribution over layers for each visual patch; (iii) \molshort$_{hybrid}$ that measures the disagreement between the \molshort$_{layer}$ and \molshort$_{patch}$ and assigns gating weights to the reserve layer (\S\ref{sec:preliminary}) representations, which subsequently are pushed through weighted linear combination to form the final patch embedding sequence. We leave out some arrows (e.g., text query to LLM backbone) for brevity.}
    \label{fig:method}
\end{figure*}


\subsection{Preliminary}
\label{sec:preliminary}
We propose \molfull~(\molshort), an instruction-aligned layer routing mechanism that dynamically selects the most useful vision layers for each input image and text instruction. \molshort~can be seamlessly attached to different MLLM backbones without changing the language model architecture. 
Suppose we have aligned vision and text encoders as in dual-encoder architectures like CLIP \citep{radford2021clip}, SigLIP  \citep{zhai2023sigmoid}, or DINOv2 w/ Txt \cite{jose2025dinov2}. Define the vision encoder's hidden states for its $L$ layers as:
\begin{equation}
\mathbf{H}^{1}, \mathbf{H}^{2}, \dots, \mathbf{H}^{L}, \qquad \mathbf{H}^{\ell} \in \mathbb{R}^{N \times d_v},
\end{equation}
where $N$ is the number of visual tokens and $d_v$ is the vision encoder hidden dimension. Let the vision encoder's corresponding text encoder\footnote{We refer to vision tower-aligned text encoders as in CLIP \citep{radford2021clip}, SigLIP \citep{zhai2023sigmoid}, or DINOv2 w/ Txt \cite{jose2025dinov2}} path produce the prompt embedding sequence
\begin{equation}
\mathbf{T} = [\mathbf{t}_1, \dots, \mathbf{t}_M] \in \mathbb{R}^{M \times d_t},
\quad
\bar{\mathbf{t}} = \mathrm{Pool}(\mathbf{T}) \in \mathbb{R}^{d_t}
\end{equation}
where $\mathbf{T}$ is a sequence of $M$ embedded text tokens of dimension $d_t$ and $\mathrm{Pool}$ is an arbitrary pooling operator.
We use the pooled text representation $\bar{\mathbf{t}}$ as a query to score
vision-layer features for routing:
\begin{equation}
\mathbf{q}=W_Q\bar{\mathbf{t}}, \qquad
\mathbf{k}^{\ell}_n=W_K\mathbf{h}^{\ell}_n, \qquad
s^{(p)}_{n,\ell}=
\frac{\mathbf{q}^{\top}\mathbf{k}^{\ell}_n}{\sqrt{d_v}} .
\end{equation}
where $\mathbf{h}^{\ell}_n$ is the hidden state for each layer $\ell$ and patch token $n$. We pass these hidden states into our routing mechanism, which treats each vision layer as an \textit{expert}, similar to MoE \citep{shazeer2017}. Among the $L$ layers of the pre-trained vision encoder, we optionally designate a ``reserve layer'' $r \in \{1,\dots,L\}$ as a stable fallback when routing is uncertain or patch-level evidence is noisy; these layers are the last or penultimate layers originally selected by the backbone MLLMs. Reserve layers, as we show in Section \ref{sec:ablations}, provide an essential anchor representation to enhance visual reasoning in \molshort{}. For variants with a reserve mechanism, the router selects the top-$k$ non-reserve layers $\mathcal{S}_n$ and combines them with a reserve layer $r$ through a learned gate:
\begin{equation}
\tilde{\mathbf{h}}_n =
(1-\gamma_n)\sum_{\ell \in \mathcal{S}_n}
\hat{q}_{n,\ell}\mathbf{h}^{\ell}_n
+
\gamma_n \mathbf{h}^{r}_n,
\qquad
\hat{q}_{n,\ell}
=
\frac{q_{n,\ell}}{\sum_{j\in\mathcal{S}_n}q_{n,j}} .
\end{equation}
Here, $\gamma_n \in [0,1]$ is the learned reserve-layer gate for patch $n$, and $q_{n,\ell}$ denotes the router distribution over non-reserve layers. 
This prevents the model from collapsing to potentially erroneous layers while still allowing it to exploit specialized intermediate representations when appropriate. We denote the set of non-reserve layers by $\mathcal{L}_{\mathrm{nonres}} = \{1,\dots,L\} \setminus \{r\}$. The final routing distribution, therefore, is decomposed into
reserve and non-reserve mass:
\(p_{n,r} = \gamma_n, \qquad p_{n,\ell} = (1-\gamma_n) q_{n,\ell}, \quad \ell \in \mathcal{L}_{\mathrm{nonres}},\)
We subsequently train the model with the standard autoregressive loss, $\mathcal{L}_{\mathrm{LM}}$, and an auxiliary load-balancing loss, $\mathcal{L}_{\mathrm{lb}}$, adopted from Switch transformers~\citep{fedus2022switch}, to prevent router collapse by encouraging balanced usage of vision encoder layers; $\lambda_{\mathrm{lb}}$ controls its strength.:
\(\mathcal{L} = \mathcal{L}_{\mathrm{LM}} + \lambda_{\mathrm{lb}} \mathcal{L}_{\mathrm{lb}}\)
We further elaborate on the training details in the Appendix.

\subsection{\molfull}

There are three different ways to calculate the image features using our approach:

\paragraph{Layer-level Sampling (\molshort$_{layer}$)}
The simplest variant performs one shared routing decision for the whole image. We first summarize each layer by average-pooling its visual tokens $\bar{\mathbf{h}}^{\ell} = \frac{1}{N}\sum_{n=1}^{N}\mathbf{h}^{\ell}_n$. Using the text query $\bar{\mathbf{t}}$, the router computes a layer distribution $p_{\ell}$ over $\ell \in \{1,\dots,L\}$, selects the top-$k$ layers $\mathcal{S}=\mathrm{TopK}(\mathbf{p},k)$, and renormalizes their probabilities as $\hat{p}_{\ell}=p_{\ell}/\sum_{j\in\mathcal{S}}p_j$. The routed image representation is then written as:
\(\tilde{\mathbf{H}} = \sum_{\ell\in\mathcal{S}}\hat{p}_{\ell}\mathbf{H}^{\ell}\)
Thus, \molshort$_{layer}$ applies the same selected layers and weights to all patches in the image.

\paragraph{Patch-level Sampling (\molshort$_{patch}$)}

Patch-level routing instead makes a separate layer selection decision for each visual token. For patch $n$, the router computes a text-conditioned distribution $p_{n,\ell}$ over layers, selects $\mathcal{S}_n=\mathrm{TopK}(\mathbf{p}_{n,:},k)$, and renormalizes the selected weights as $\hat{p}_{n,\ell}=p_{n,\ell}/\sum_{j\in\mathcal{S}_n}p_{n,j}$. The routed patch representation is 
\(\tilde{\mathbf{h}}_n=\sum_{\ell\in\mathcal{S}_n}\hat{p}_{n,\ell}\mathbf{h}^{\ell}_n\)

This allows different visual patch tokens (i.e., image regions) to draw from different vision-layer abstractions while keeping the number of output patch tokens unchanged.

\paragraph{Hybrid Sampling (\molshort$_{hybrid}$)}
\molshort$_{hybrid}$ unifies layer-level and patch-level sampling to combine the stability of global routing with the flexibility of local routing, improving adaptability across downstream tasks. We first compute a global routing distribution $p^{(g)}_{\ell}$ and patch-level, local routing distribution $p^{(p)}_{n,\ell}$ as
\begin{equation}
p^{(g)}_{\ell} = \mathrm{softmax}\left(\frac{s^{(g)}_{\ell}}{\tau}\right),
\quad
p^{(p)}_{n,\ell} = \mathrm{softmax}_{\ell}\left(\frac{s^{(p)}_{n,\ell}}{\tau}\right).
\end{equation}
Here, $p^{(g)}_{\ell}$ is the image-level routing probability for layer $\ell$, and $p^{(p)}_{n,\ell}$ is the patch-level routing probability for patch $n$ and layer $\ell$. 
We then predict a text prompt--dependent mixing weight:
\(\beta = \sigma(\mathbf{w}_{\beta}^{\top}\bar{\mathbf{t}} + b_{\beta}).\)
$\beta \in [0,1]$ controls the interpolation between global and local routing, $\mathbf{w}_{\beta}$ and $b_{\beta}$ are learned parameters, and $\sigma(\cdot)$ is the sigmoid function. The hybrid non-reserve routing distribution is computed via a weighted product-of-experts \citep{hinton1999products}:
\(q_{n,\ell} \propto \left(p^{(g)}_{\ell}\right)^{1-\beta} \left(p^{(p)}_{n,\ell}\right)^{\beta}.\)
Here, $q_{n,\ell}$ is the fused non-reserve routing probability for patch $n$ and layer $\ell$. Equivalently, in log-space, this would be written as follows:
\begin{equation}
\log q_{n,\ell} = (1-\beta)\log p^{(g)}_{\ell} + \beta \log p^{(p)}_{n,\ell} - \log Z_n,
\end{equation}
where $Z_n$ is the normalization constant for patch $n$.

As discussed above (\S\ref{sec:preliminary}), we also use a \textit{reserve layer}, $r$, a penultimate or last layer of the vision tower, as a fallback layer. To determine \textit{when} to fallback or not, we measure the disagreement between local and global routing with
\begin{equation}
d_n = \mathrm{KL}\left(p^{(p)}_{n,:} \,\|\, p^{(g)}\right)
= \sum_{\ell \neq r} p^{(p)}_{n,\ell}
\left(\log p^{(p)}_{n,\ell} - \log p^{(g)}_{\ell}\right).
\end{equation}
where $d_n$ is the disagreement score for patch $n$, $\mathrm{KL}(\cdot\|\cdot)$ denotes KL divergence, and $r$ is the reserve layer index. Then, we map disagreement to a reserve layer probability as $\gamma_n = \sigma(w_d d_n + b_d)$ wherein $\gamma_n \in [0,1]$ is the reserve weight for patch $n$, and $w_d, b_d$ are learned scalar parameters. Intuitively, the higher the $d_{n}$, i.e., the disagreement between the layer- and patch-level distributions, the more we rely on the reserve layer and vice versa. Letting $\mathcal{S}_n = \mathrm{TopK}(\mathbf{q}_{n,:}, k)$ denote the selected non-reserve layers, the final patch representation is
\begin{equation}
\tilde{\mathbf{h}}_n =
(1-\gamma_n)\sum_{\ell \in \mathcal{S}_n} \hat{q}_{n,\ell}\mathbf{h}^{\ell}_n
+ \gamma_n \mathbf{h}^{r}_n,
\quad
\text{where}
\quad
\hat{q}_{n,\ell} = \frac{q_{n,\ell}}{\sum_{j \in \mathcal{S}_n} q_{n,j}}.
\end{equation}
Here, $\hat{q}_{n,\ell}$ denotes the renormalized hybrid routing weight, and $\mathbf{h}^{r}_n$ is the representation of patch $n$ from the reserve layer $r$.

\paragraph{Sampling from Layers of Multiple Vision Encoders} 
Unlike prior works that concatenate or interleave last-layer features from multiple vision encoders \citep{Tong_2024_CVPR, tong2024cambrian}, \molshort~can first route within each vision
encoder and then fuse the routed representations across vision encoders. For each vision encoder $m$, we apply
one of the proposed routing variants, $\alpha_m \in \{\mathrm{layer}, \mathrm{patch},
\mathrm{hybrid}\}$, to obtain
\begin{equation}
\tilde{\mathbf{H}}^{(m)}
=
\mathrm{MoL}^{(m)}_{\alpha_m}
\left(
\{\mathbf{H}^{(m,\ell)}\}_{\ell=1}^{L_m},
\bar{\mathbf{t}}^{(m)}
\right),
\end{equation}
where $\bar{\mathbf{t}}^{(m)}$ is the vision encoder-aligned text representation. For instance, in the CLIP + DINOv2 w/ Txt setting in Table~\ref{tab:single_vs_multi_encoder}, we apply \molshort$_{hybrid}$ independently to both vision encoders and fuse their routed representations through concatenation and upsampling. After routing, we project both branches into the language model's hidden space:
\(\mathbf{Z}^{c} = \Pi_c(\tilde{\mathbf{H}}^{c}), \qquad \mathbf{Z}^{d} = \Pi_d(\tilde{\mathbf{H}}^{d}),\)
where $c$ and $d$ denote CLIP and DINOv2, respectively, and $\Pi_c$ and $\Pi_d$ are the corresponding vision-language projectors. Here, $\mathbf{Z}^{c} \in \mathbb{R}^{N_c \times d_{\mathrm{LLM}}}$ and $\mathbf{Z}^{d} \in \mathbb{R}^{N_d \times d_{\mathrm{LLM}}}$, where $N_c$ and $N_d$ are the numbers of patch tokens from each vision encoder, and $d_{\mathrm{LLM}}$ is the language model's hidden dimension. 
Since DINOv2 and CLIP branches produce 256 and 576 patch tokens, respectively, we first upsample the projected DINOv2 features to the CLIP token-grid size:
\(\hat{\mathbf{Z}}^{d} = \mathrm{Upsample}_{N_d \rightarrow N_c}(\mathbf{Z}^{d}) \in \mathbb{R}^{N_c \times d_{\mathrm{LLM}}},\)
where $\mathrm{Upsample}$ uses bilinear interpolation. We subsequently concatenate the two representations along the feature dimension and pass them through a learned linear projector:
\begin{equation}
\mathbf{F}
= \left[\mathbf{Z}^{c} \,\Vert\, \hat{\mathbf{Z}}^{d}\right]\mathbf{W}_{f}
+ \mathbf{1}_{N_c}\mathbf{b}_{f}^{\top},
\qquad
\mathbf{F} \in \mathbb{R}^{N_c \times d_{\mathrm{LLM}}},
\end{equation}
where $\Vert$ denotes feature concatenation, $\mathbf{W}_{f} \in \mathbb{R}^{2d_{\mathrm{LLM}} \times d_{\mathrm{LLM}}}$ and $\mathbf{b}_{f} \in \mathbb{R}^{d_{\mathrm{LLM}}}$ are learned parameters, and $\mathbf{1}_{N_c}$ broadcasts the bias across patch tokens. This fusion allows the model to combine query-relevant representations from both vision encoders while preserving the CLIP branch's output token length. The resulting patch embedding sequence $\mathbf{F}$ is then passed to the language model.

\section{Experiments}
\label{sec:experiments}


\begin{table*}[t]
    \caption{\textbf{Performance across backbones and router types on fine-grained visual reasoning benchmarks.}
    Metrics reported are accuracy. \textit{V$^\ast$} benchmark is broken into its constituent sub-categories for detailed analysis. 
    \textbf{Best $\Delta$} denotes the maximum improvement (in absolute accuracy points) achieved by any \molshort{} variant over the corresponding backbone baseline for each metric. We include \molshort$_\text{layer}$ for the Vicuna-13B settings with CLIP and DINOv2 w/ Txt as global-routing comparisons. The multi-encoder results are reported separately in Table~\ref{tab:single_vs_multi_encoder}.}
    \label{tab:mol_backbone_router_perf}
    \centering
    \setlength{\tabcolsep}{5pt}
    \resizebox{\textwidth}{!}{
        \begin{tabular}{l *{11}{c}}
            \toprule
            \textbf{Method} &
            \multicolumn{5}{c}{\textit{\textbf{V$^\ast$}}} &
            \multirow{2}{*}{\textbf{MMStar}} &
            \multirow{2}{*}{\textbf{HRBench8K}} &
            \multirow{2}{*}{\textbf{HRBench4K}} &
            \multirow{2}{*}{\textbf{RealWorldQA}} &
            \multirow{2}{*}{\textbf{NaturalBench}} &
            \multirow{2}{*}{\textbf{CharXiv}} \\
            & \textit{all} & \textit{GPT4V-hard} & \textit{OCR} & \textit{direct attr.} & \textit{rel. pos.} & & & & & & \\
            \midrule

            \rowcolor{groupgray}
            \multicolumn{12}{c}{\textit{LLaVA-v1.5-13B (Vicuna-13B + CLIP)}~\citep{liu2024llavav1.5, radford2021clip}} \\
            LLaVA-v1.5-13B  & 71.43 & 52.94 & 50.00 & 65.22 & 93.42 & 30.00 & 36.88 & 43.75 & 51.64 & 68.93 & 23.84 \\
            MoL$_{\text{layer}}$  & 76.89 & 58.82 & 53.33 & 75.65 & 92.11 & 29.53 & 38.00 & 40.62 & 52.35 & 69.71 & 26.25 \\
            MoL$_{\text{patch}}$  & \textbf{84.87} & 58.82 & \textbf{60.00} & \textbf{86.09} & \textbf{98.68} & 30.13 & 38.50 & 42.25 & 51.82 & 69.41 & \textbf{29.17} \\
            MoL$_{\text{hybrid}}$ & 84.03 & \textbf{64.71} & 56.67 & 85.22 & 97.37 & \textbf{31.07} & \textbf{38.62} & \textbf{43.88} & \textbf{53.14} & \textbf{70.36} & 25.56 \\
            
            \rowcolor{avgblue}
            \textcolor{mygreen}{\textbf{Best $\Delta$}} 
            & \textcolor{mygreen}{\textbf{+13.44}}
            & \textcolor{mygreen}{\textbf{+11.77}}
            & \textcolor{mygreen}{\textbf{+10.00}}
            & \textcolor{mygreen}{\textbf{+20.87}}
            & \textcolor{mygreen}{\textbf{+5.26}}
            & \textcolor{mygreen}{\textbf{+1.07}}
            & \textcolor{mygreen}{\textbf{+1.74}}
            & \textcolor{mygreen}{\textbf{+0.13}}
            & \textcolor{mygreen}{\textbf{+1.50}}
            & \textcolor{mygreen}{\textbf{+1.43}}
            & \textcolor{mygreen}{\textbf{+5.33}} \\
            \midrule

            \rowcolor{groupgray}
            \multicolumn{12}{c}{\textit{Vicuna-13B + DINOv2 w/ Txt}~\citep{liu2024llavav1.5, jose2025dinov2}} \\
            Vicuna-13B + DINOv2  & 65.55 & \textbf{76.47} & 40.00 & 67.83 & 69.74 & 28.87 & 32.75 & 35.38 & 50.46 & \textbf{68.29} & 22.92 \\
            MoL$_{\text{layer}}$  & 67.65 & 52.94 & 36.67 & 60.87 & 93.42 & 27.80 & 34.12 & 35.38 & 50.46 & 67.92 & 25.94 \\
            MoL$_{\text{patch}}$  & 76.05 & 64.71 & 40.00 & 74.78 & 94.74 & 28.47 & 33.37 & 34.88 & 51.02 & 66.03 & 17.71 \\
            MoL$_{\text{hybrid}}$ & \textbf{84.45} & 70.59 & \textbf{50.00} & \textbf{86.09} & \textbf{98.68} & \textbf{30.13} & \textbf{36.50} & \textbf{39.87} & \textbf{52.29} & 67.89 & \textbf{26.92} \\
            
            \rowcolor{avgblue}
            \textcolor{mygreen}{\textbf{Best $\Delta$}} 
            & \textcolor{mygreen}{\textbf{+18.90}} & \textcolor{myred}{-5.88} & \textcolor{mygreen}{\textbf{+10.00}} & \textcolor{mygreen}{\textbf{+18.26}} & \textcolor{mygreen}{\textbf{+28.94}}
            & \textcolor{mygreen}{\textbf{+1.26}} & \textcolor{mygreen}{\textbf{+3.75}} & \textcolor{mygreen}{\textbf{+4.49}} & \textcolor{mygreen}{\textbf{+1.83}} & \textcolor{myred}{-0.40} & \textcolor{mygreen}{\textbf{+4.00}} \\
            \midrule

            \rowcolor{groupgray}
            \multicolumn{12}{c}{\textit{Llama-3-8B-V (LLama-3-8B-Instruct + SigLIP)}~\citep{grattafiori2024llama, zhai2023sigmoid}} \\
            Llama-3-8B-V  & 54.62 & 47.06 & 66.67 & 56.52 & 48.68 & \textbf{37.73} & \textbf{44.00} & \textbf{51.12} & \textbf{59.87} & 74.75 & 25.05 \\
            MoL$_{\text{patch}}$  & 52.52 & 41.18 & 43.33 & 60.00 & 47.37 & 37.07 & 40.00 & 48.88 & 58.92 & \textbf{75.02} & 26.16 \\
            MoL$_{\text{hybrid}}$ & \textbf{63.03} & \textbf{47.06} & \textbf{76.67} & \textbf{70.43} & \textbf{50.00} & 33.40 & 42.13 & 45.00 & 45.62 & 72.05 & \textbf{26.82} \\
            
            \rowcolor{avgblue}
            \textcolor{mygreen}{\textbf{Best $\Delta$}} 
            & \textcolor{mygreen}{\textbf{+8.42}} & \textcolor{mygreen}{+0.00} & \textcolor{mygreen}{\textbf{+10.00}} & \textcolor{mygreen}{\textbf{+13.91}} & \textcolor{mygreen}{\textbf{+1.32}}
            & \textcolor{myred}{-0.66} & \textcolor{myred}{-1.87} & \textcolor{myred}{-2.24} & \textcolor{myred}{-0.95} & \textcolor{mygreen}{\textbf{+0.27}} & \textcolor{mygreen}{\textbf{+1.77}} \\
            \midrule

            \rowcolor{groupgray}
            \multicolumn{12}{c}{\textit{Phi-1.5-1.3B + SigLIP}~\citep{li2023textbooks, zhai2023sigmoid}} \\
            Phi-1.5-1.3B  & 42.01 & 23.52 & 26.66 & 35.65 & 61.84 & 34.93 & 32.00 & 36.50 & 34.93 & 67.31 & 4.07 \\
            MoL$_{\text{patch}}$  & 39.92 & 29.41 & 50.00 & 28.70 & 55.26 & 33.60 & 30.63 & 33.12 & 38.82 & 60.86 & 19.65 \\
            MoL$_{\text{hybrid}}$ & \textbf{50.84} & \textbf{23.52} & \textbf{43.33} & \textbf{42.60} & \textbf{72.36} & 34.93 & 32.00 & \textbf{37.25} & \textbf{39.74} & \textbf{68.00} & \textbf{20.46} \\
            \rowcolor{avgblue}
            \textcolor{mygreen}{\textbf{Best $\Delta$}} 
            & \textcolor{mygreen}{\textbf{+8.83}} 
            & \textcolor{mygreen}{\textbf{+0.00}} 
            & \textcolor{mygreen}{\textbf{+16.67}} 
            & \textcolor{mygreen}{\textbf{+6.95}} 
            & \textcolor{mygreen}{\textbf{+10.52}}
            & \textcolor{mygreen}{+0.00} 
            & \textcolor{mygreen}{+0.00} 
            & \textcolor{mygreen}{\textbf{+0.75}} 
            & \textcolor{mygreen}{\textbf{+4.81}} 
            & \textcolor{mygreen}{\textbf{+0.69}} 
            & \textcolor{mygreen}{\textbf{+16.39}} \\
            \midrule
        \end{tabular}
    }
\end{table*}

\begin{table*}[t]
    \caption{\textbf{Performance comparison against existing state-of-the-art methods on multi-layer visual feature integration.} With all the existing baselines providing LLaVA-baseline checkpoints, we use LLaVA-v1.5-13B~\citep{liu2024llavav1.5} as our baseline in this experiment.}
    \label{tab:mol_baseline_router_perf}
    \centering
    \setlength{\tabcolsep}{5pt}
    \resizebox{\textwidth}{!}{
        \begin{tabular}{l *{11}{c}}
            \toprule
            \textbf{Method} &
            \multicolumn{5}{c}{\textit{\textbf{V$^\ast$}}} &
            \multirow{2}{*}{\textbf{MMStar}} &
            \multirow{2}{*}{\textbf{HRBench8K}} &
            \multirow{2}{*}{\textbf{HRBench4K}} &
            \multirow{2}{*}{\textbf{RealWorldQA}} &
            \multirow{2}{*}{\textbf{NaturalBench}} &
            \multirow{2}{*}{\textbf{CharXiv}} \\
            
            & \textit{all} & \textit{GPT4V-hard} & \textit{OCR} & \textit{direct attr.} & \textit{rel. pos.} & & & & & & \\
            \midrule

            LLaVA-v1.5-13B
            & 71.43 & 52.94 & 50.00 & 65.22 & 93.42 & 30.00 & 36.88 & 43.75 & 51.64 & 68.93 & 23.84 \\

            TokenPacker \citep{li2025tokenpacker}
            & 73.53 & 47.06 & \textbf{63.33} & 66.09 & 93.74 & 29.07 & 38.00 & 43.38 & 54.12 & 67.63 & 22.76  \\

            MLVF \citep{lin2025multi}
            & 78.92 & 45.88 & 54.00 & 79.13 & 96.05 & 29.84 & 36.88 & 42.50 & 51.90 & 69.80 & 24.72  \\

            DeepStack-L \citep{meng2024deepstack}
            & 65.13 & 64.71 & 46.67 & 56.52 & 85.53 & 29.26 & 37.63 & 43.75 & \textbf{57.77} & 69.10 & 21.31 \\

            Dense Connector \citep{yao2024dense}
            & 81.35 & 47.05 & 56.67 & 83.47 & \textbf{98.68} & 30.06 & 37.12 & 43.75 & 52.41 & \textbf{71.41} & 25.23 \\

            IGVA \citep{lin2025multi}
            & 70.38 & 45.12 & 52.20 & 66.79 & 88.60 & 29.26 & 36.17 & 42.10 & 51.85 & 67.63 & 23.89 \\

            TGIF \citep{lin2026text}
            & 63.87 & 52.94 & 50.00 & 56.52 & 82.89 & 29.07 & 33.75 & 40.63 & 54.38 & 69.20 & 20.00 \\

            \midrule
            
            \rowcolor{avgblue}
            \textbf{MoL$_{\text{layer}}$}
            & 76.89 & 58.82 & 53.33 & 75.65 & 92.11 & 29.53 & 38.00 & 40.62 & 52.35 & 69.71 & 27.85 \\

            \rowcolor{avgblue}
            \textbf{MoL$_{\text{patch}}$}
            & \textbf{84.87} & 58.82 & 60.00 & \textbf{86.09} & \textbf{98.68} & 30.13 & 38.50 & 42.25 & 51.82 & 69.41 & \textbf{28.65} \\

            \rowcolor{avgblue}
            \textbf{MoL$_{\text{hybrid}}$}
            & 84.03 & \textbf{64.71} & 56.67 & 85.22 & 97.37 & \textbf{30.47} & \textbf{38.62} & \textbf{43.88} & 53.14 & 70.36 & 26.97 \\

            \bottomrule
        \end{tabular}
    }
\end{table*}


\begin{table*}[t]
    \caption{\textbf{Comparison of single- and multi-encoder routing with a Vicuna-13B backbone.}
    Interleaved-MoF interleaves CLIP and DINOv2 patch inputs. \molshort{} routes within one or both encoders; the dual-encoder variant uses concatenation/upsampling followed by a learned projector. \textbf{Best $\Delta$} reports the largest improvement among the three \molshort{} configurations over Interleaved-MoF, computed from the displayed scores in percentage points. $\dagger$ denotes DINOv2 with an aligned text encoder.}
    \label{tab:single_vs_multi_encoder}
    \centering
    \setlength{\tabcolsep}{5pt}
    \resizebox{\textwidth}{!}{
        \begin{tabular}{l l *{11}{c}}
            \toprule
            \textbf{Method} & \textbf{Visual Input} &
            \multicolumn{5}{c}{\textit{\textbf{V$^\ast$}}} &
            \multirow{2}{*}{\textbf{MMStar}} &
            \multirow{2}{*}{\textbf{HRBench8K}} &
            \multirow{2}{*}{\textbf{HRBench4K}} &
            \multirow{2}{*}{\textbf{RealWorldQA}} &
            \multirow{2}{*}{\textbf{NaturalBench}} &
            \multirow{2}{*}{\textbf{CharXiv}} \\
            & & \textit{all} & \textit{GPT4V-hard} & \textit{OCR} & \textit{direct attr.} & \textit{rel. pos.} & & & & & & \\
            \midrule

            Interleaved-MoF~\citep{Tong_2024_CVPR} & CLIP + DINOv2 & 82.35 & 58.82 & 53.33 & 82.61 & \textbf{98.68} & 31.87 & 35.75 & \textbf{44.12} & \textbf{57.77} & 67.26 & 26.73 \\

             Dual Enc No Router & CLIP + DINOv2 & 84.45 & 70.59 & 66.67 & 85.22 & 93.42 & 30.47 & 36.13 & 43.50 & 53.86 & 69.12 & 28.97 \\

            \molshort$_{\text{hybrid}}$ & CLIP & 84.03 & 64.71 & 56.67 & 85.22 & 97.37 & 31.07 & \textbf{38.62} & 43.88 & 53.14 & \textbf{70.36} & 25.56 \\

            \molshort$_{\text{hybrid}}$ & DINOv2$^{\dagger}$ & 84.45 & 70.59 & 50.00 & 86.09 & \textbf{98.68} & 30.13 & 36.50 & 39.87 & 52.29 & 67.89 & 26.92 \\

            \molshort$_{\text{hybrid}}$ - Dual Enc. & CLIP + DINOv2$^{\dagger}$ & \textbf{90.34} & \textbf{82.35} & \textbf{66.67} & \textbf{92.17} & \textbf{98.68} & \textbf{32.00} & 37.62 & 43.50 & 52.55 & 69.72 & \textbf{31.13} \\

            \rowcolor{avgblue}
            \textbf{Best $\Delta$} & & \textcolor{mygreen}{\textbf{+7.99}} & \textcolor{mygreen}{\textbf{+23.53}} & \textcolor{mygreen}{\textbf{+13.34}} & \textcolor{mygreen}{\textbf{+9.56}} & \textcolor{mygreen}{\textbf{+0.00}} & \textcolor{mygreen}{\textbf{+0.13}} & \textcolor{mygreen}{\textbf{+2.87}} & \textcolor{myred}{\textbf{-0.24}} & \textcolor{myred}{\textbf{-4.63}} & \textcolor{mygreen}{\textbf{+3.10}} & \textcolor{mygreen}{\textbf{+4.40}} \\
            \bottomrule
        \end{tabular}
    }
\end{table*}


In this section, we elaborate on the experimental setup of the backbone multimodal LLMs (MLLMs) we apply \molfull~(\molshort{}) to, discuss the downstream fine-grained visual understanding and reasoning benchmarks, detail the analytical frameworks employed to study layer-wise sampling behaviors, and demonstrate how \molshort~sampled layer representations enhance fine-grained visual understanding in MLLMs.
\paragraph{Models} Our work builds \molshort~in a fully reproducible MLLM training framework built on top of \citep{liu2024llavav1.5, he2024bunny}. This framework provides both the vision-language connector pre-training and multimodal instruction-tuning stages, unlike open-weight-only models \citep{bai2025qwen3, grattafiori2024llama} where the underlying training pipeline data is not fully available. Namely, in this work we use LLaVA-1.5-13B \citep{liu2024llavav1.5}, Llama-3-8B-V \citep{grattafiori2024llama}, and Phi-1.5-1.3B\citep{li2023textbooks} + SigLIP \citep{zhai2023sigmoid} as experimental baselines, all of which are open-sourced by the Bunny framework \citep{he2024bunny}. The pre-trained vision encoders span CLIP \citep{radford2021clip}, DINOv2 w/ Txt\citep{oquabdinov2}, and SigLIP \citep{zhai2023sigmoid}. For DINOv2 w/ Txt we pair it with Vicuna-13B, which is equivalent of a LLaVA-v1.5-13B with DINOv2 w/ Txt as the vision encoder. For additional training details, please refer to Appendix.

\paragraph{Dataset} We adopt the exact pre-training and instruction-tuning datasets provided by the individual backbone MLLMs \citep{liu2024llavav1.5, deitke2025molmo, he2024bunny} to ensure a fair comparison. For fine-grained visual understanding benchmarks, we employ \textit{V*} \citep{wu2024v}, HRBench4K, HRBench8K \citep{wang2025divide}, NaturalBench \citep{li2024naturalbench}, RealWorldQA \citep{xai-realworldqa}, and CharXiv \citep{wang2024charxiv} to evaluate the baseline and \molshort-enhanced capabilities of different MLLMs. 

\subsection{Effect of \molfull~on Model Perception}

\paragraph{Fine-grained Visual Understanding}
Table~\ref{tab:mol_backbone_router_perf} shows that \molshort{} improves fine-grained visual reasoning across multiple backbone architectures and vision encoders, across different parameter sizes (13B, 8B, 1.5B). On the primary LLaVA-v1.5-13B setting, \molshort{} yields consistent gains over the baseline, with especially large improvements on \textit{V*} overall (+13.44), direct attribute recognition (+20.87), OCR (+10.00), and CharXiv (+5.33). Similar gains appear with DINOv2 w/ Txt, where \molshort$_{\text{hybrid}}$ improves \textit{V*} by +18.90 and relative position by +28.94, suggesting that adaptive layer routing is not specific to CLIP-based vision encoders. The gains are also visible in smaller or different backbones: for Phi-1.5-1.3B, \molshort{} substantially improves OCR (+16.67), relative position (+10.52), and CharXiv (+16.39). Across settings, \molshort$_{\text{patch}}$ is often strongest on localized perceptual categories, while \molshort$_{\text{hybrid}}$ tends to perform better on benchmarks requiring both local grounding and broader reasoning. These results suggest that selecting task-relevant intermediate layers is a general mechanism for improving fine-grained perception, rather than a backbone-specific optimization.

\paragraph{Comparison against State-of-the-Art}
In Table~\ref{tab:mol_baseline_router_perf}, we compare \molshort~with prior approaches that aggregate intermediate visual features via multi-resolution stacking (DeepStack~\citep{meng2024deepstack}), token compression and fusion (TokenPacker~\citep{li2025tokenpacker}), or dense cross-layer connections (MLVF~\citep{lin2025multi}, Dense Connector~\citep{yao2024dense}). Unlike these fixed aggregation schemes, \molshort~adaptively routes vision-layer representations conditioned on the input instruction.
\molshort~consistently improves fine-grained visual reasoning. \molshort$_{\text{patch}}$ achieves the best overall \textit{V*} performance, outperforming Dense Connector by 3.52 points, with particularly strong gains on localized categories such as \texttt{OCR} and \texttt{direct attribute}. \molshort$_{\text{hybrid}}$ further improves performance on more challenging benchmarks (e.g., GPT4V-hard, MMStar, HRBench), indicating the benefit of combining global and patch-level routing. Importantly, \molshort~operates with a single vision encoder yet matches or outperforms prior multi-encoder approaches (e.g., Interleaved-MoF \citep{Tong_2024_CVPR}) as shown in Table \ref{tab:single_vs_multi_encoder}, demonstrating that adaptive layer routing can be more effective than increasing visual diversity through simple interleaving of additional encoder features or more patch tokens.

\noindent\begin{minipage}[t]{0.52\linewidth}
\vspace{0pt}
\paragraph{Coarse-grained Visual Understanding}
Although \molshort~targets fine-grained perception, we also evaluate standard VQA benchmarks in Table~\ref{tab:mol_coarse_grained_perf}. \molshort~largely preserves coarse-grained reasoning performance, with small gains on POPE, GQA, and MMMU, and minimal change on TextVQA. We include the full results here and defer additional analysis to the Appendix.
\end{minipage}\hfill
\begin{minipage}[t]{0.45\linewidth}
    \vspace{0pt}
    \captionof{table}{\textbf{Coarse-grained VQA performance.}}
    \label{tab:mol_coarse_grained_perf}
    \centering
    \scriptsize
    \setlength{\tabcolsep}{2pt}
    \renewcommand{\arraystretch}{0.78}
    \begin{tabular}{@{}lcccc@{}}
        \toprule
        \textbf{Model} & \textbf{POPE} & \textbf{GQA} & \textbf{TextVQA} & \textbf{MMMU} \\
        \midrule
    
        \rowcolor{groupgray}
        \multicolumn{5}{c}{\textit{LLaVA-v1.5-13B (Vicuna-13B + CLIP)}~\citep{liu2024llavav1.5, radford2021clip}} \\
        LLaVA-v1.5-13B  & 85.60 & 63.26 & \textbf{61.25} & 36.89 \\
        MoL$_{\text{patch}}$  & 86.23 & 63.73 & 61.09 & \textbf{38.22} \\
        MoL$_{\text{hybrid}}$ & \textbf{86.47} & \textbf{63.75} & 61.11 & 37.00 \\
    
        \rowcolor{avgblue}
        \textcolor{mygreen}{\textbf{Best $\Delta$}} &
        \textcolor{mygreen}{\textbf{+0.87}} &
        \textcolor{mygreen}{\textbf{+0.49}} &
        \textcolor{myred}{-0.14} &
        \textcolor{mygreen}{\textbf{+1.33}} \\
        \midrule
    
        \rowcolor{groupgray}
        \multicolumn{5}{c}{\textit{Vicuna-13B + DINOv2 w/ Txt}~\citep{liu2024llavav1.5, jose2025dinov2}} \\
        Vicuna-13B + DINOv2  & 84.05 & 61.46 & 46.87 & 34.67 \\
        MoL$_{\text{patch}}$  & \textbf{84.12} & 59.96 & \textbf{47.12} & 34.11 \\
        MoL$_{\text{hybrid}}$ & 84.08 & 60.57 & 46.80 & 34.67 \\
    
        \rowcolor{avgblue}
        \textcolor{mygreen}{\textbf{Best $\Delta$}} &
        \textcolor{mygreen}{\textbf{+0.07}} &
        \textcolor{myred}{-0.89} &
        \textcolor{mygreen}{+0.25} &
        \textcolor{mygreen}{+0.00} \\
        \bottomrule
    \end{tabular}
\end{minipage}
\par\medskip

\subsection{\molfull~with Multiple Vision Encoders}
\label{sec:multi_encoder}
\paragraph{Setup}
We extend hybrid routing to a Vicuna-13B backbone with CLIP and DINOv2 w/ Txt, applying a separate text-conditioned router to each vision encoder. The evaluated configuration selects one non-reserve layer per patch ($k=1$) and combines it with the encoder's reserve representation using the disagreement gate. Each routed stream is projected to the language model's hidden dimension. The concatenation-upsampling fusion resizes the DINOv2 stream to the CLIP token-grid size, concatenates the two streams along the feature dimension, and applies a learned linear projection to produce 576 visual tokens. Table~\ref{tab:single_vs_multi_encoder} compares this configuration with Interleaved-MoF~\citep{Tong_2024_CVPR}, and an additional baseline for dual encoder representation concatenation-upsampling without the router (Dual Enc. No Router), where both build on Vicuna-13B backbone. This is a comparison between complete systems; the pretrained encoders and fusion mechanisms are not a controlled routing-only ablation.

\paragraph{Results}
The multi-encoder configuration achieves 90.34 on \textit{V*}, compared with 82.35 for Interleaved-MoF. The largest gain is on GPT4V-hard (82.35 versus 58.82), followed by OCR (66.67 versus 53.33) and direct attributes (92.17 versus 82.61); relative-position accuracy is tied at 98.68. These changes correspond to 19 additional correct answers out of 238: four on GPT4V-hard, four on OCR, and eleven on direct attributes. The same checkpoint achieves 31.13 on CharXiv. The remaining reported comparisons are mixed, including lower HRBench4K and RealWorldQA scores. We therefore do not infer a uniform advantage across benchmarks from this comparison.

\paragraph{Interpretation and Insights}
A plausible explanation is that routing and encoder diversity address different sources of lost visual information. Routing exposes query-relevant intermediate representations within each encoder, while the two encoders supply different learned representations of the image. The learned channel projection gives the language model access to both routed streams without extending the sequence beyond the CLIP token count. Combining an image-level routing preference with patch-specific evidence also allows the selected abstraction to vary across regions, while the reserve representation provides an anchor. The concentration of gains in OCR, attributes, and challenging visual questions is consistent with this explanation; the unchanged relative-position result suggests that extra visual diversity is less useful when the baseline is already near ceiling.

\subsection{Analysis of Sampling Behavior} 
\label{sec:analysis_sampling_behavior}

\paragraph{Layer-wise Sampling Behavior}

\begin{figure*}[t]
    \centering
    \includegraphics[width=\linewidth]{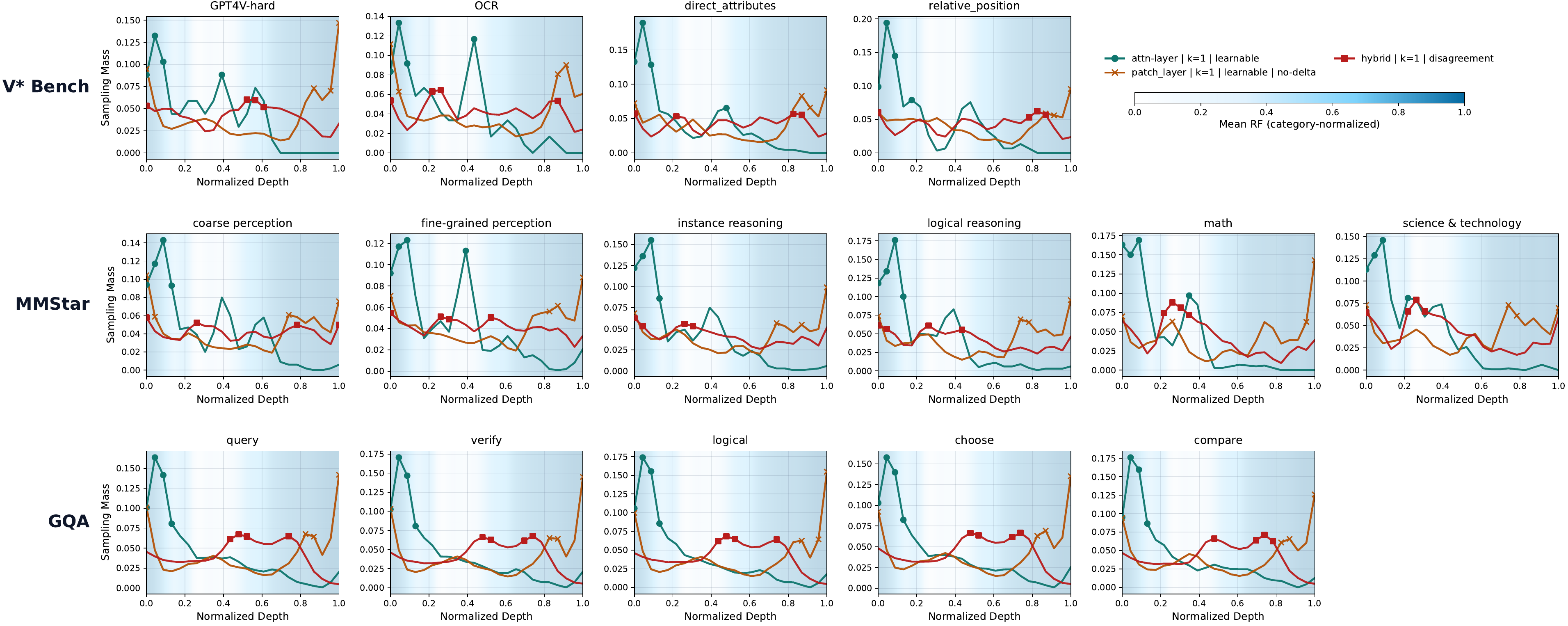}
    \caption{\textbf{Layer selection behaviors of \molfull.}
    Average normalized routing probability across vision-encoder layers for LLaVA-v1.5-13B on \textit{V*}Bench, MMStar, and GQA categories. Background intensity shows category-normalized receptive field (RF)~\citep{park2023what}; lighter regions indicate more localized layers. Markers denote selected top-$k$ layers ($k=4$). DINOv2 variants are shown in Appendix Figure~\ref{fig:app_vstar_mmstar_gqa_rf}.}
    \label{fig:layer_selection_behavior}
\end{figure*}

Figure \ref{fig:layer_selection_behavior} shows that \molshort~learns non-uniform, task-dependent routing patterns across benchmarks and question categories. The layer-level router (\molshort$_{layer}$), which assigns one distribution to the entire image consistently prioritizes early layers, especially for categories requiring localized evidence such as OCR, direct attributes, and relative position. This is because global routing must rely on features that are consistently informative across the whole image, and early layers preserve fine-grained spatial details before they are aggregated. By contrast, the patch-level router (\molshort$_{patch}$) places more mass on later layers. Since it computes routing independently for each patch, it can leverage semantically enriched representations that encode broader context, allowing individual regions to be interpreted relative to the overall scene and query. As a result, later-layer selections often reflect patches that benefit from contextualized, higher-level features. The hybrid router (\molshort$_{hybrid}$) balances these two behaviors by combining global and patch-level routing, yielding more consistent mid-to-late layer usage. Overall, these patterns suggest that global routing favors detail-preserving features, while patch-level routing can exploit contextualized representations, motivating the need for both.

\paragraph{Receptive Field Analysis across Different Layers}
To gain a deeper understanding of the routing behavior in light of the above layer-wise sampling behiavor analysis, we analyze the receptive field (RF) of each layer using the average self-attention distance~\citep{dosovitskiy2021an}:
$
\mathrm{RF}^{\ell} = \mathbb{E}_{i}\left[\sum_{j} A^{\ell}_{i,j} d(i,j)\right].
$
RF increases with depth: early layers capture local interactions, while later layers aggregate broader spatial context. This provides a mechanistic explanation for the routing patterns. \molshort$_{\text{layer}}$ favors early, small-RF layers because global routing must rely on detail-preserving features that are consistently informative across the image. In contrast, \molshort$_{\text{patch}}$ often selects later, large-RF layers, since patch-wise routing can leverage contextualized representations to interpret each region relative to the scene and query. \molshort$_{\text{hybrid}}$ combines both regimes, enabling access to both local evidence and global context. Thus, RF analysis complements the routing observations by showing that \molshort{} adapts to the inherent locality-context trade-off encoded across vision layers.

\begin{figure*}[t]
    \centering
    \includegraphics[width=\linewidth]{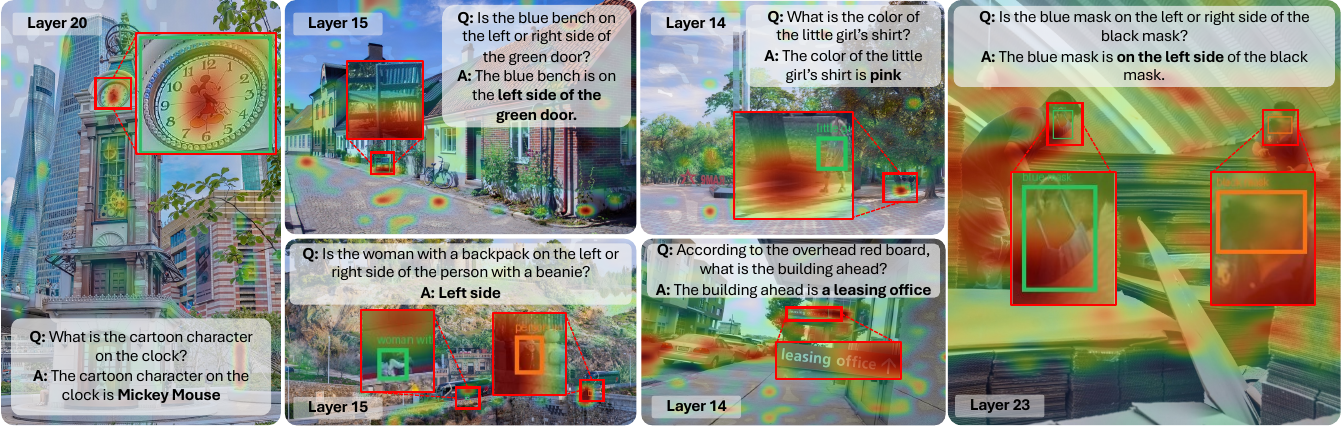}
    \caption{\textbf{Self-attention heatmap visualization of the selected layers aggregated across all the queries of the layers on the selected layers of CLIP-ViT-L14.} For additional qualitative analysis on the selected layers of the DINOv2 encoder, please refer to Figure \ref{fig:app_case_study_attention_heatmap_dinov2}.
    } 
    \label{fig:case_study_attention_heatmap}
\end{figure*}

\begin{table}[t]
    \caption{\textbf{Ablation on vision encoder-aligned text encoder.} 
    \molshort$_{image}$ uses image-only routing, where the pooled embeddings of each vision encoder layer is converted to logits to form a distribution over the layers, while \molshort$_{hybrid}$ uses text-conditioned routing. We denote the improvement / degradation compared to the baseline model in {\color{mygreen}green} / {\color{myred}red}. We use a Vicuna-13B + CLIP backbone.}
    \label{tab:ablation_no_text_encoder}
    \centering
    \scriptsize
    \setlength{\tabcolsep}{3pt}
    \renewcommand{\arraystretch}{0.78}
    \begin{tabular}{@{}lcccc@{}}
        \toprule
        \textbf{Model} & \textbf{\textit{V*}} & \textbf{GQA} & \textbf{TextVQA} & \textbf{MMMU} \\
        \midrule
        \molshort$_{\text{image}}$ 
        & 77.31 {\color{mygreen}(+5.88)} 
        & 62.94 {\color{myred}(-0.32)} 
        & 58.17 {\color{myred}(-3.08)} 
        & 36.00 {\color{myred}(-0.89)} \\

        \molshort$_{\text{hybrid}}$ 
        & \textbf{84.03} {\color{mygreen}\textbf{(+12.60)}}
        & \textbf{63.75} {\color{mygreen}\textbf{(+0.49)}}
        & \textbf{61.11} {\color{myred}(-0.14)}
        & \textbf{37.00} {\color{mygreen}\textbf{(+0.11)}}\\

        \bottomrule
    \end{tabular}
\end{table}

\begin{table*}[t]
    \caption{\textbf{Ablations on routing design.}
    (a) Removing the reserve layer degrades performance, indicating its importance as a fallback when routing is uncertain. In tandem, having an anchor (\textit{reserve}) layer with complementary $k$ layers helps MLLMs perform grounded fine-grained visual reasoning.
    (b) Varying $k$ shows that selecting a small set of layers suffices for strong fine-grained reasoning.}
    \label{tab:ablation_combined_k_and_reserve_layer}
    \centering
    \begin{subtable}[t]{0.46\textwidth}
        \centering
        \resizebox{0.7\linewidth}{!}{%
    \begin{tabular}{lrr}
        \toprule
        \textbf{Benchmark} & \molshort$_\text{hybrid}$ & \shortstack{\molshort$_\text{hybrid}$\\\rule{0.4\linewidth}{0.3pt}\\No Reserve} \\
        \midrule
        \textit{V$^\ast$} & 82.35 & 78.57 \textcolor{myred}{(-3.78\%)} \\
        HRBench8K & 38.62 & 35.50 \textcolor{myred}{(-3.12\%)} \\
        HRBench4K & 43.88 & 38.75 \textcolor{myred}{(-5.13\%)} \\
        POPE & 86.50 & 82.10 \textcolor{myred}{(-4.40\%)} \\
        MMMU & 37.00 & 34.00 \textcolor{myred}{(-3.00\%)} \\
        \bottomrule
    \end{tabular}%
        }
        \caption{Effect of removing the reserve layer.}
        \label{tab:reserve_ablation}
    \end{subtable}
    \hfill
    \begin{subtable}[t]{0.53\textwidth}
        \centering
        \resizebox{\linewidth}{!}{%
        \begin{tabular}{lrrrrrrrr}
            \toprule
            \textbf{Benchmark} & \multicolumn{4}{c}{\molshort$_\text{patch}$} & \multicolumn{4}{c}{\molshort$_\text{hybrid}$} \\
            \cmidrule(lr){2-5}\cmidrule(lr){6-9}
            & $k=1$ & $k=2$ & $k=4$ & Best & $k=1$ & $k=2$ & $k=4$ & Best \\
            \midrule
            \textit{V$^\ast$} & 76.89 & 73.95 & 79.41 & 79.41 & 84.03 & 76.89 & 67.65 & 84.03 \\
            HRBench8K & 39.38 & 36.50 & 37.25 & 39.38 & 38.62 & 38.25 & 34.88 & 38.62 \\
            HRBench4K & 43.12 & 42.75 & 43.75 & 43.75 & 43.88 & 44.00 & 43.50 & 44.00 \\
            MMStar & 30.13 & 31.93 & 29.93 & 31.93 & 30.47 & 30.00 & 29.73 & 30.47 \\
            NaturalBench & 69.41 & 68.07 & 68.61 & 69.41 & 69.96 & 68.91 & 70.45 & 70.45 \\
            CharXiv & 28.65 & 28.47 & 27.94 & 28.65 & 26.13 & 26.25 & 26.97 & 26.97 \\
            RealWorldQA & 55.42 & 54.90 & 55.16 & 55.42 & 55.03 & 56.60 & 56.34 & 56.60 \\
            \bottomrule
        \end{tabular}%
        }
        \caption{Effect of $k$ on fine-grained visual reasoning.}
        \label{tab:k_ablation}
    \end{subtable}
\end{table*}

\subsection{Ablations} 
\label{sec:ablations}

\paragraph{Role of Vision Encoder-Aligned Text Encoders}
Table \ref{tab:ablation_no_text_encoder} studies whether routing should be conditioned on the text encoder aligned with the vision encoder. We compare \molshort$_{image}$, which predicts layer-routing probabilities from vision hidden states alone, against the text-conditioned \molshort$_{hybrid}$. Vision-only routing substantially underperforms the baseline, especially on \textit{V*}, where it drops by 9.16 points. We hypothesize that image-only routing lacks a task-specific signal for determining which layer representations are relevant, making the router more prone to selecting inconsistent or semantically mismatched layers during joint optimization. In contrast, \molshort$_{hybrid}$ improves \textit{V*} by 13.44 points and yields small gains on GQA and MMMU while largely preserving TextVQA performance. These results suggest that layer selection should not be treated as an image-only decision: the input instruction provides an essential signal for selecting which visual abstractions are relevant for the current task.

\paragraph{Role of Reserve Layers and the Number of Layers} 
Table~\ref{tab:ablation_combined_k_and_reserve_layer} highlights two key properties of \molshort{}. First, removing the reserve layer consistently degrades performance (3-5 points across benchmarks), indicating that it acts as a stable \textit{anchor} that preserves strong, text-aligned representations when routing over intermediate layers is uncertain or noisy. Second, varying $k$ shows that performance does not benefit from aggregating many layers; instead, small $k$ values (e.g., $k=1$ or $k=4$) are sufficient and often optimal. Together, these results suggest that effective fine-grained visual reasoning arises from a balance between stability and selectivity: the reserve layer provides robustness, while sparse routing over a few task-relevant layers avoids introducing noise from irrelevant representations.


\paragraph{Qualitative Analysis on the Self-Attention Layers of Vision Encoders}
Figure~\ref{fig:case_study_attention_heatmap} shows that \molshort~selects layers whose self-attention maps align with query-relevant regions. For localized tasks such as OCR and attribute recognition, mid-to-early layers (< layers 14-15) attend to fine-grained details (e.g., text on signs, small objects), while later layers highlight broader semantic regions (e.g., people, objects). For spatial reasoning queries, attention is distributed across multiple relevant regions, enabling the model to resolve relationships such as left/right positioning.
We defer \textbf{Limitations and Broader Impacts} to Appendix \ref{app:limitations_broader_impacts} due to lack of space.

\section{Conclusion}
\label{sec:conclusion}

We introduced \molfull~(\molshort), an instruction-guided layer routing framework for MLLMs that dynamically selects intermediate vision-encoder representations for fine-grained visual reasoning conditioned on the input textual instruction. Instead of relying on fixed final-layer features or increasing the visual token budget, \molshort~selects task-relevant patch representations through layer-level, patch-level, and hybrid routing for both a single and dual vision encoder schemes. Across fine-grained visual understanding benchmarks, \molshort~consistently improves over backbone MLLMs and prior static feature-integration methods, while remaining competitive on conventional coarse-grained VQA tasks. Our analysis further shows that different routing strategies exploit different regions of the vision-encoder hierarchy: layer-level routing favors localized early layers, patch-level routing leverages contextualized later layers, and hybrid routing effectively combines both regimes to reap the benefit of both worlds. Overall, \molshort~demonstrates that dynamically selecting the appropriate visual abstraction can be a more effective alternative to scaling visual inputs or adding multiple vision encoders for improving MLLM perception and reasoning.

\section{Acknowledgements}
\label{sec:acknowledgement}
This research is based upon work supported by DARPA ITM Program No. FA8650-23-C-7316, the Office of the Director of National Intelligence (ODNI), Intelligence Advanced Research Projects Activity (IARPA), via 560000C260018, CapitalOne-Illinois Center for Generative AI Safety, Knowledge Systems, and Cybersecurity (ASKS), Amazon-Illinois Center on AI for Interactive Conversational Experiences (AICE). The views and conclusions contained herein are those of the authors and should not be interpreted as necessarily representing the official policies, either expressed or implied, of DARPA, IARPA, ODNI, DARPA, or the U.S. Government. The U.S. Government is authorized to reproduce and distribute reprints for governmental purposes notwithstanding any copyright annotation therein. This research used the Delta and DeltaAI advanced computing and data resources, which are supported by the National Science Foundation (award OAC 2320345 and award OAC 2005572) and the State of Illinois. Delta and DeltaAI are joint efforts of the University of Illinois Urbana-Champaign and its National Center for Supercomputing Applications. The AI Research Institutes program by National Science Foundation and the Institute of Education Sciences, U.S. Department of Education through Award 2229873 - AI Institute for Transforming Education for Children with Speech and Language Processing Challenges.

{
    \bibliographystyle{plain}
    \bibliography{references}
}

\clearpage
\appendix
\phantomsection
\section*{\centering Appendix}
\label{app:appendix_guide}

\noindent\textbf{How to read this appendix.}
Section~\ref{app:limitations_broader_impacts} states the scope, limitations, and broader impacts of the work. Sections~\ref{app:training_details}--\ref{app:evaluation_details} document how the models and evaluations were run. Sections~\ref{app:additional_results}--\ref{app:qualitative_examples} provide the extended empirical evidence behind the main-paper results. The links in the roadmap jump directly to each section.

\vspace{1em}
\begin{center}
\renewcommand{\arraystretch}{1.35}
\begin{tabular}{p{0.24\textwidth}p{0.66\textwidth}}
    \toprule
    \textbf{Appendix Section} & \textbf{Contents} \\
    \midrule
    \textbf{\hyperref[app:limitations_broader_impacts]{A. Limitations and Broader Impacts}} &
    Scope of claims, methodological limitations, expected benefits, risks, and responsible-use considerations. \\
    \textbf{\hyperref[app:training_details]{B. Training Details}} &
    Hardware, two-stage training, optimization settings, objective, and router naming. \\
    \textbf{\hyperref[app:evaluation_details]{C. Evaluation Details}} &
    Benchmark scripts, output files, score aggregation, and missing-entry conventions. \\
    \textbf{\hyperref[app:additional_results]{D. Additional Benchmark Results}} &
    Full benchmark matrix, $k$ sweep, and reserve-layer ablation. \\
    \textbf{\hyperref[app:router_analysis]{E. Router and RF Analyses}} &
    Layer-selection behavior, receptive-field overlays, router usage diagnostics, and DINOv2 placeholders. \\
    \textbf{\hyperref[app:qualitative_examples]{F. Qualitative Examples}} &
    Selected attention panels plus a LLaVA + DINOv2 heatmap analysis. \\
    \textbf{\hyperref[app:limitations_reproducibility]{G. Reproducibility Notes}} &
    Remaining caveats, source artifacts, and checkpoint-config conventions. \\
    \bottomrule
\end{tabular}
\end{center}
\renewcommand{\arraystretch}{1.0}

\clearpage

\section{Limitations and Broader Impacts}
\label{app:limitations_broader_impacts}

\paragraph{Scope of the empirical claims.}
\molfull{} is designed to improve the visual representation interface between a pretrained vision encoder and a multimodal language model by routing over intermediate vision-layer features. The central claim is therefore not that \molshort{} universally improves every multimodal task, but that instruction-conditioned access to task-relevant intermediate representations improves fine-grained visual reasoning while largely preserving standard VQA performance. This distinction matters: the strongest gains in our experiments appear on benchmarks that require localized evidence, such as V$^\ast$Bench, CharXiv, HRBench, OCR-style questions, direct attribute recognition, and relative-position reasoning. On more conventional or coarse-grained VQA benchmarks, the gains are smaller and sometimes close to neutral. We view this as a useful boundary on the contribution rather than a weakness of the evaluation: \molshort{} is targeted at perception bottlenecks where the final vision layer can discard details needed by the language model.

\paragraph{Architectural limitations.}
\molshort{} assumes access to the hidden states of a pretrained vision encoder and, for the text-conditioned variants, a text encoder aligned with that vision tower. This makes the method natural for open and reproducible MLLM training pipelines such as LLaVA- and Bunny-style models, but less directly applicable to vision encoder-free MLLM architectures \citep{chen2024solo, cui2025emu3} or models that expose only final visual embeddings. The router also introduces additional trainable parameters, routing hyperparameters such as $k$, and a reserve-layer design choice. Our ablations show that a small number of selected layers and the reserve layer are important, but these choices may need to be retuned for substantially different vision encoders, tokenization schemes, resolutions, or downstream domains.

\paragraph{Evaluation limitations.}
The experiments cover several backbones, parameter scales, and vision encoders, including CLIP, SigLIP, and DINOv2 w/ Txt, but they do not exhaust the space of modern MLLM architectures. We do not claim that the exact best router variant transfers unchanged to every backbone: \molshort$_{\mathrm{patch}}$ tends to help localized perceptual categories, while \molshort$_{\mathrm{hybrid}}$ is more robust when both local grounding and broader context are needed. We also report single-run results for most large-scale training settings because training 8B--13B multimodal models is computationally expensive. The breadth of benchmarks and ablations partially mitigates this limitation, but future work should add repeated runs, stronger uncertainty estimates, and evaluation on additional domains such as video, medical imagery, remote sensing, and document-heavy workflows.

\paragraph{Interpretability limitations.}
Router probabilities, receptive-field overlays, and selected-layer attention maps provide useful diagnostics for understanding how \molshort{} uses intermediate representations. They should not, however, be interpreted as complete causal explanations of the model's answer. A selected layer may contribute useful features even when its attention map is diffuse, and visually plausible attention can still coincide with an incorrect answer. We therefore use these analyses as evidence about routing behavior and locality-context trade-offs, not as a guarantee that the final language-model reasoning is faithful or safe.

\paragraph{Broader positive impacts.}
Improved fine-grained visual reasoning can benefit applications that require careful visual grounding, such as assistive image understanding, chart and figure interpretation, educational tools, accessibility support, and scientific document analysis. A practical advantage of \molshort{} is that it can improve access to useful visual abstractions without necessarily increasing the number of visual tokens or requiring multiple vision encoders. This may offer a more compute-conscious path than simply scaling image resolution, concatenating additional towers, or passing substantially longer visual sequences to the language model.

\paragraph{Risks and misuse.}
The same fine-grained perception improvements can also increase risks when applied to sensitive visual data. Better OCR, attribute recognition, and small-object grounding may make it easier to extract private information from images, analyze people or locations without consent, or strengthen surveillance-style applications. \molshort{} also does not remove inherited biases from the underlying language model, vision encoder, or instruction-tuning data. Since the method improves perception rather than safety alignment, it should not be treated as sufficient for high-stakes deployment in domains such as medical diagnosis, legal decision-making, hiring, biometric identification, or safety-critical robotics.

\paragraph{Responsible-use considerations.}
We frame \molshort{} as a representation-routing method for research on multimodal perception, not as a complete deployment system. Systems built on top of this work should include privacy-aware data handling, domain-specific evaluation, bias and robustness audits, and human oversight in high-stakes settings. Releasing training details, normalized router names, checkpoint-configuration conventions, and analysis scripts is intended to make the empirical claims easier to audit and to help future work identify both successes and failure modes rather than treating improved benchmark scores as a blanket guarantee of reliability.

\section{Training Details}
\label{app:training_details}

All models were trained on a single server with four NVIDIA H100 NVL GPUs, each with 95830 MiB of device memory. We use DeepSpeed ZeRO-2 for the vision-language connector pretraining stage and DeepSpeed ZeRO-3 for the multimodal instruction-tuning stage. Unless otherwise stated, training uses bf16, tf32, gradient checkpointing, a cosine learning-rate schedule, warmup ratio 0.03, weight decay 0, maximum sequence length 2048, and lazy image preprocessing.

The training objective is the standard autoregressive language-modeling loss with an auxiliary router load-balancing loss:
\begin{equation}
\mathcal{L} = \mathcal{L}_{\mathrm{LM}} + \lambda_{\mathrm{lb}}\mathcal{L}_{\mathrm{lb}}.
\end{equation}
The default load-balancing coefficient is $\lambda_{\mathrm{lb}}=10^{-2}$. During training, the router temperature is linearly annealed from 2.0 to 1.0. We save router statistics during training for the analysis in Appendix~\ref{app:router_analysis}.

\paragraph{Router naming.}
We group the \texttt{attn-token} implementation variants as \molshort$_{\mathrm{layer}}$, and report the better-performing layer-level variant for each setting. The \texttt{patch\_layer} implementation corresponds to \molshort$_{\mathrm{patch}}$, and the \texttt{hybrid} implementation corresponds to \molshort$_{\mathrm{hybrid}}$.

Table~\ref{tab:app_training_setup} summarizes the hardware and optimization recipe used for each model family. The main distinction is between the Vicuna/LLaVA training path, which follows the LLaVA-v1.5 data mixture, and the Bunny training path, which uses the Bunny pretraining and instruction-tuning data for the Llama-3 and Phi backbones. We report global batch sizes after accounting for the four GPUs and gradient accumulation.

\begin{table*}[t]
    \centering
    \small
    \setlength{\tabcolsep}{3pt}
    \resizebox{\textwidth}{!}{
    \begin{tabular}{llcccccc}
        \toprule
        \textbf{Backbone} & \textbf{Stage} & \textbf{Data} & \textbf{ZeRO} & \textbf{Epochs} & \textbf{Global Batch} & \textbf{Projector LR} & \textbf{Router LR} \\
        \midrule
        Vicuna-13B + CLIP & Pretrain & BLIP/LAION-CC-SBU 558K & 2 & 1 & 256 & $1{\times}10^{-3}$ & $1{\times}10^{-5}$--$5{\times}10^{-5}$ \\
        Vicuna-13B + CLIP & Finetune & LLaVA-v1.5 mix 665K & 3 & 1 & 128 & $2{\times}10^{-5}$ & $2{\times}10^{-5}$ \\
        Llama-3-8B + SigLIP & Pretrain & Bunny pretrain 2M & 2 & 1 & 128 & $1{\times}10^{-3}$ & $1{\times}10^{-4}$ \\
        Llama-3-8B + SigLIP & Finetune & Bunny instruction 695K & 3 & 1 & 128 & $2{\times}10^{-5}$--$2{\times}10^{-4}$ & $2{\times}10^{-5}$--$2{\times}10^{-4}$ \\
        Phi-1.5 + SigLIP & Pretrain & Bunny pretrain 2M & 2 & 1 & 256 & $5{\times}10^{-4}$ & $5{\times}10^{-4}$ \\
        Phi-1.5 + SigLIP & Finetune & Bunny instruction 695K & 3 & 1 & 128 & $5{\times}10^{-5}$--$2{\times}10^{-4}$ & $5{\times}10^{-5}$--$2{\times}10^{-4}$ \\
        \bottomrule
    \end{tabular}
    }
    \caption{\textbf{Training setup.} The Vicuna-13B + CLIP setting follows the LLaVA-v1.5 two-stage training recipe. The Llama-3-8B + SigLIP and Phi-1.5 + SigLIP settings follow the Bunny two-stage recipe and use LoRA during instruction tuning.}
    \label{tab:app_training_setup}
\end{table*}

Table~\ref{tab:app_router_config} gives the naming bridge between implementation keys in checkpoints and the method names used throughout the paper. This is important because the experiments include multiple layer-level router implementations. We treat them as the same conceptual family, \molshort$_{\mathrm{layer}}$, when summarizing broad benchmark results.

\begin{table*}[t]
    \centering
    \small
    \setlength{\tabcolsep}{4pt}
    \resizebox{\textwidth}{!}{
    \begin{tabular}{lcccccc}
        \toprule
        \textbf{Variant} & \textbf{Implementation key} & \textbf{Routing granularity} & \textbf{$k$} & \textbf{Reserve layer} & \textbf{Reserve mode} & \textbf{Text conditioning} \\
        \midrule
        \molshort$_{\mathrm{layer}}$ & \texttt{attn-token} & image-level layer set & 1--4 & $-2$ & local or scaled gate & attention \\
        \molshort$_{\mathrm{patch}}$ & \texttt{patch\_layer} & patch-specific layer set & 1--4 & $-2$ & local gate & attention \\
        \molshort$_{\mathrm{hybrid}}$ & \texttt{hybrid} & global-local fused routing & 1--4 & $-2$ & disagreement gate & attention \\
        \bottomrule
    \end{tabular}
    }
    \caption{\textbf{Router configuration summary.} The reserve layer is the penultimate vision-encoder layer used as an anchor representation. Appendix tables use normalized method names rather than implementation keys.}
    \label{tab:app_router_config}
\end{table*}

\section{Evaluation Details}
\label{app:evaluation_details}

We evaluate with benchmark-specific scripts under \texttt{scripts/v1\_5/eval}. Each run writes predictions, per-example details, and a summary file under \texttt{playground/data/eval/BENCHMARK/answers}. Unless the benchmark defines a specialized metric, we report accuracy. For V$^\ast$Bench, HRBench4K/8K, MMStar, NaturalBench, RealWorldQA, CharXiv, and the conventional VQA benchmarks, scores are percentages. For benchmarks with multiple splits or categories, the reported number is the official aggregate used by the corresponding evaluation script.

The broad benchmark matrix in Appendix~\ref{app:additional_results} is generated from \texttt{playground/data/eval/benchmark\_overall\_table.md}. Empty entries indicate that the run was not available in the generated table. We omit columns that were entirely empty for the selected LLaVA-v1.5 family rows.

\section{Additional Benchmark Results}
\label{app:additional_results}

Tables~\ref{tab:app_full_benchmark_1} and~\ref{tab:app_full_benchmark_2} expand the main-paper results to a broader set of benchmarks for the primary LLaVA-v1.5 family. The table is split only for readability. These numbers are intended to show whether \molshort{} preserves general multimodal ability while improving fine-grained visual reasoning. The strongest gains concentrate on fine-grained benchmarks such as V$^\ast$Bench, CharXiv, HRBench, and MMStar, while conventional VQA-style benchmarks usually remain close to the baseline.

\begin{table*}[t]
    \centering
    \scriptsize
    \setlength{\tabcolsep}{3pt}
    \resizebox{\textwidth}{!}{
    \begin{tabular}{lrrrrrrrrrrrr}
        \toprule
        \textbf{Method} & \textbf{MMBench} & \textbf{MMMU} & \textbf{MMStar} & \textbf{MMVet} & \textbf{MMVP} & \textbf{CV-Bench} & \textbf{NaturalBench} & \textbf{CharXiv} & \textbf{CountBenchQA} & \textbf{GQA} & \textbf{POPE} & \textbf{TextVQA} \\
        \midrule
        LLaVA-v1.5-13B & 76.49 & 36.89 & 30.00 & 41.11 & 65.33 & 62.60 & 68.93 & 25.20 & 50.71 & 63.26 & 85.60 & 61.25 \\
        \molshort$_{\mathrm{layer}}$ & 75.81 & 36.56 & 29.53 & -- & 64.67 & 57.88 & 69.71 & 27.85 & 48.68 & 63.43 & 86.53 & 60.64 \\
        \molshort$_{\mathrm{patch}}$ & 76.40 & 38.22 & 31.93 & -- & 65.00 & 63.28 & 69.41 & 30.89 & 48.68 & 63.73 & 86.23 & 61.09 \\
        \molshort$_{\mathrm{hybrid}}$ & 75.94 & 37.00 & 31.07 & 41.88 & 65.67 & 62.46 & 70.45 & 26.97 & 51.93 & 63.75 & 86.47 & 61.11 \\
        \bottomrule
    \end{tabular}
    }
    \caption{\textbf{Full benchmark matrix for the primary LLaVA-v1.5 family, part 1.} \molshort$_{\mathrm{layer}}$ is the per-metric best of the \texttt{attn-layer} and \texttt{attn-token} layer-level implementations. \molshort$_{\mathrm{patch}}$ and \molshort$_{\mathrm{hybrid}}$ are the best available scores within their normalized method families.}
    \label{tab:app_full_benchmark_1}
\end{table*}

\begin{table*}[t]
    \centering
    \scriptsize
    \setlength{\tabcolsep}{4pt}
    \resizebox{\textwidth}{!}{
    \begin{tabular}{lrrrrrrrrr}
        \toprule
        \textbf{Method} & \textbf{DocVQA} & \textbf{MME-RW} & \textbf{HallusionBench} & \textbf{RealWorldQA} & \textbf{ChartQA} & \textbf{HRBench4K} & \textbf{HRBench8K} & \textbf{V$^\ast$Bench} & \textbf{DTD} \\
        \midrule
        LLaVA-v1.5-13B & 23.98 & 32.33 & 51.64 & 55.29 & 20.28 & 43.75 & 36.88 & 71.43 & 35.41 \\
        \molshort$_{\mathrm{layer}}$ & 24.68 & 30.39 & 55.09 & 55.82 & 20.56 & 41.38 & 38.00 & 76.89 & 31.94 \\
        \molshort$_{\mathrm{patch}}$ & 25.28 & 33.18 & 52.97 & 55.42 & 21.08 & 43.75 & 39.38 & 84.87 & -- \\
        \molshort$_{\mathrm{hybrid}}$ & 25.21 & 32.34 & 53.23 & 56.60 & 22.44 & 44.00 & 38.62 & 84.03 & -- \\
        \bottomrule
    \end{tabular}
    }
    \caption{\textbf{Full benchmark matrix for the primary LLaVA-v1.5 family, part 2.} MME-RW denotes MME-RealWorld. Empty entries indicate unavailable scores in the generated benchmark table.}
    \label{tab:app_full_benchmark_2}
\end{table*}

Table~\ref{tab:app_fine_grained_k_sweep} isolates the effect of the number of selected layers. The main pattern is that selecting more layers is not automatically better. A small $k$ often gives the best result, which supports the interpretation that the router should selectively retrieve a few useful abstraction levels instead of averaging many layers indiscriminately.

\begin{table*}[t]
    \centering
    \scriptsize
    \setlength{\tabcolsep}{4pt}
    \resizebox{\textwidth}{!}{
    \begin{tabular}{lrrrrrrrr}
        \toprule
        \textbf{Benchmark} & \multicolumn{4}{c}{\molshort$_{\mathrm{patch}}$} & \multicolumn{4}{c}{\molshort$_{\mathrm{hybrid}}$} \\
        \cmidrule(lr){2-5}\cmidrule(lr){6-9}
        & $k=1$ & $k=2$ & $k=4$ & Best & $k=1$ & $k=2$ & $k=4$ & Best \\
        \midrule
        V$^\ast$Bench & 76.89 & 73.95 & 79.41 & 79.41 & 84.03 & 76.89 & 67.65 & 84.03 \\
        HRBench8K & 39.38 & 36.50 & 37.25 & 39.38 & 38.62 & 38.25 & 34.88 & 38.62 \\
        HRBench4K & 43.12 & 42.75 & 43.75 & 43.75 & 43.88 & 44.00 & 43.50 & 44.00 \\
        MMStar & 30.13 & 31.93 & 29.93 & 31.93 & 30.47 & 30.00 & 29.73 & 30.47 \\
        NaturalBench & 69.41 & 68.07 & 68.61 & 69.41 & 69.96 & 68.91 & 70.45 & 70.45 \\
        CharXiv & 28.65 & 28.47 & 27.94 & 28.65 & 26.13 & 26.25 & 26.97 & 26.97 \\
        RealWorldQA & 55.42 & 54.90 & 55.16 & 55.42 & 55.03 & 56.60 & 56.34 & 56.60 \\
        \bottomrule
    \end{tabular}
    }
    \caption{\textbf{Fine-grained visual reasoning under different $k$ values.} A small selected set of layers is usually sufficient; the best $k$ varies by task and routing family.}
    \label{tab:app_fine_grained_k_sweep}
\end{table*}

Table~\ref{tab:app_reserve_ablation} tests the reserve layer directly. Removing it hurts V$^\ast$Bench, HRBench, POPE, and MMMU, indicating that the reserve representation acts as a stable fallback when the routed intermediate layers are noisy or uncertain.

\begin{table}[t]
    \centering
    \small
    \setlength{\tabcolsep}{5pt}
    \begin{tabular}{lrr}
        \toprule
        \textbf{Benchmark} & \molshort$_{\mathrm{hybrid}}$ & \shortstack{\molshort$_{\mathrm{hybrid}}$\\No Reserve} \\
        \midrule
        V$^\ast$Bench & 82.35 & 78.57 \textcolor{myred}{(-3.78)} \\
        HRBench8K & 38.62 & 35.50 \textcolor{myred}{(-3.12)} \\
        HRBench4K & 43.88 & 38.75 \textcolor{myred}{(-5.13)} \\
        POPE & 86.50 & 82.10 \textcolor{myred}{(-4.40)} \\
        MMMU & 37.00 & 34.00 \textcolor{myred}{(-3.00)} \\
        \bottomrule
    \end{tabular}
    \caption{\textbf{Reserve-layer ablation.} Removing the reserve layer consistently hurts performance, supporting its role as a stable anchor representation.}
    \label{tab:app_reserve_ablation}
\end{table}

\section{Additional Router and Receptive Field Analyses}
\label{app:router_analysis}

\begin{figure*}[th!]
    \centering
    \includegraphics[width=\linewidth]{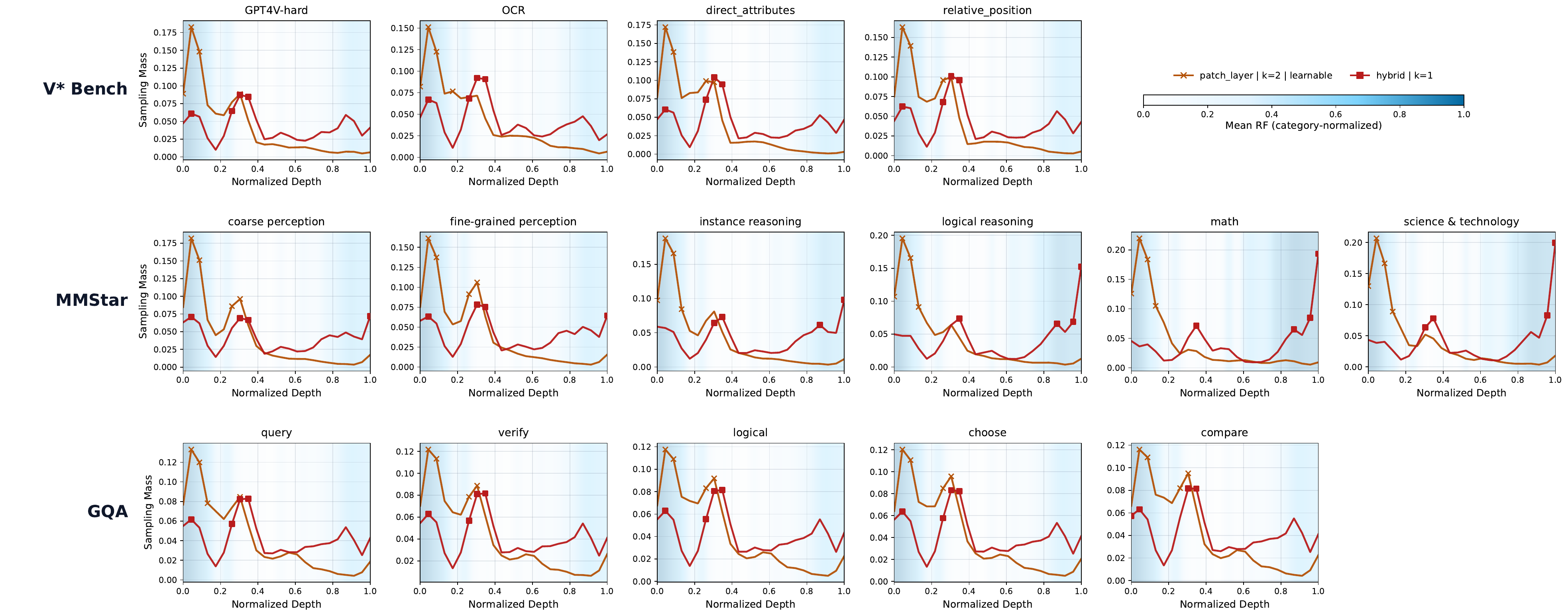}
    \caption{\textbf{DINOv2 layer sampling with receptive-field background on V$^\ast$Bench, MMStar, and GQA.}
    Curves show the normalized routing probability assigned to each DINOv2 vision-encoder layer for \molshort$_{\mathrm{patch}}$ and \molshort$_{\mathrm{hybrid}}$ across question categories. Background intensity denotes the category-normalized receptive field (RF), where lighter regions indicate more localized layers and darker regions indicate broader spatial context. Type-wise markers indicate the selected top-$k$ layers.}
    \label{fig:app_vstar_mmstar_gqa_rf}
\end{figure*}

\paragraph{DINOv2 RF-background}
Figure~\ref{fig:app_vstar_mmstar_gqa_rf} shows that DINOv2 exhibits category-dependent routing patterns consistent with the CLIP analysis in the main paper. \molshort$_{\mathrm{patch}}$ often assigns high probability to early layers, especially for V$^\ast$Bench and perceptual MMStar/GQA categories, suggesting that localized DINOv2 features remain useful for fine-grained cues such as OCR, attributes, and relative position. In contrast, \molshort$_{\mathrm{hybrid}}$ places more mass on mid-to-late layers for categories requiring broader reasoning, such as logical reasoning, math, and science/technology in MMStar. This supports the role of hybrid routing: it preserves access to local evidence while selectively relying on deeper, larger-RF layers when the query requires more semantic or relational context.

\section{Additional Qualitative Examples}
\label{app:qualitative_examples}

\begin{figure*}[t]
    \centering
    \includegraphics[width=\linewidth]{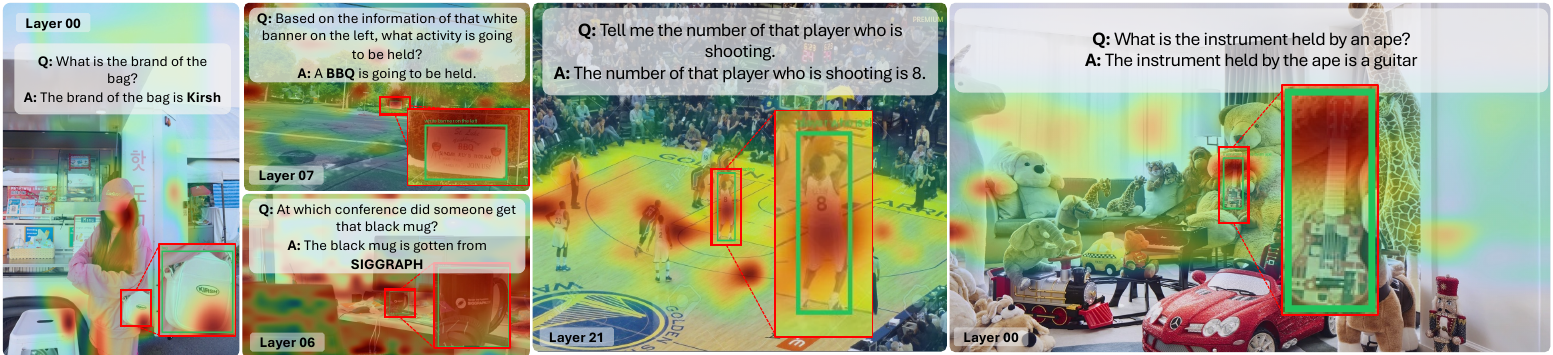}
    \caption{\textbf{Self-attention heatmap visualization of the selected layers aggregated across all the queries of the layers for DINOv2 encoder for the selected layers.}
    } 
    \label{fig:app_case_study_attention_heatmap_dinov2}
\end{figure*}

 \paragraph{Qualitative Analysis on the DINOv2 Attention Maps}
Figure~\ref{fig:app_case_study_attention_heatmap_dinov2} shows that \molshort{} with a DINOv2 encoder consistently attends to localized, query-relevant regions for fine-grained visual reasoning. For OCR-heavy examples (e.g., reading the conference name on a mug or identifying text on a banner), the selected layers concentrate attention tightly around the relevant textual regions, despite cluttered backgrounds and small font sizes. In counting and attribute-centric examples, such as identifying the jersey number of a basketball player or recognizing the brand logo on a bag, the router selects layers whose attention maps isolate the discriminative object regions while suppressing surrounding distractions. Interestingly, even in semantically broader scenes (e.g., identifying the instrument held by the ape), the selected layers maintain strong localized grounding on the queried object rather than diffusing attention across the full scene. These patterns suggest that \molshort{} with DINOv2 learns to exploit intermediate-layer representations that preserve both fine-grained local structure and semantically meaningful object boundaries, enabling grounded visual reasoning without requiring additional visual tokens or multiple encoders.

\section{Reproducibility Notes}
\label{app:limitations_reproducibility}

The appendix reports the generated result artifacts available under \texttt{playground/data/eval}. Because the main experiments train large MLLMs, we do not report multiple random seeds for every model and benchmark. We instead prioritize breadth across backbones, vision encoders, routing variants, and benchmarks. Checkpoint \texttt{config.json} files are treated as the source of truth for router mode, $k$, reserve-layer settings, and learning-rate metadata when run-script names and final checkpoint labels differ.

For reproducibility, final paper figures used by this appendix are copied into \texttt{paper\_draft\_latex/figures/appendix}. The source analysis artifacts remain in \texttt{playground/data/eval}, including the benchmark matrix, router/RF plots, reserve comparisons, and resolution analyses.

\clearpage
\section*{NeurIPS Paper Checklist}

\begin{enumerate}

\item {\bf Claims}
    \item[] Question: Do the main claims made in the abstract and introduction accurately reflect the paper's contributions and scope?
    \item[] Answer: \answerYes{}.
    \item[] Justification: The abstract and introduction state the paper's main contribution, \molfull{}, and scope the claims around fine-grained visual reasoning. The reported claims are supported by the experimental results in Section~\ref{sec:experiments}.

\item {\bf Limitations}
    \item[] Question: Does the paper discuss the limitations of the work performed by the authors?
    \item[] Answer: \answerYes{}.
    \item[] Justification: We discuss limitations in the Appendix, including the empirical scope of the evaluation and remaining directions for extending \molfull{} to broader backbones, modalities, and settings.
    \item[] Guidelines:
    \begin{itemize}
        \item The answer \answerNA{} means that the paper has no limitation while the answer \answerNo{} means that the paper has limitations, but those are not discussed in the paper. 
        \item The authors are encouraged to create a separate ``Limitations'' section in their paper.
        \item The paper should point out any strong assumptions and how robust the results are to violations of these assumptions.
        \item The authors should reflect on the scope of the claims made.
        \item The authors should reflect on the factors that influence the performance of the approach.
        \item The authors should discuss the computational efficiency of the proposed algorithms and how they scale with dataset size.
        \item If applicable, the authors should discuss possible limitations of their approach to address problems of privacy and fairness.
        \item Reviewers will be specifically instructed to not penalize honesty concerning limitations.
    \end{itemize}

\item {\bf Theory assumptions and proofs}
    \item[] Question: For each theoretical result, does the paper provide the full set of assumptions and a complete (and correct) proof?
    \item[] Answer: \answerNA{}.
    \item[] Justification: This paper does not present theoretical results or formal proofs.
    \item[] Guidelines:
    \begin{itemize}
        \item The answer \answerNA{} means that the paper does not include theoretical results. 
        \item All the theorems, formulas, and proofs in the paper should be numbered and cross-referenced.
        \item All assumptions should be clearly stated or referenced in the statement of any theorems.
        \item The proofs can either appear in the main paper or the supplemental material.
        \item Any informal proof provided in the core of the paper should be complemented by formal proofs.
        \item Theorems and Lemmas that the proof relies upon should be properly referenced. 
    \end{itemize}

\item {\bf Experimental result reproducibility}
    \item[] Question: Does the paper fully disclose all the information needed to reproduce the main experimental results of the paper to the extent that it affects the main claims and/or conclusions of the paper?
    \item[] Answer: \answerYes{}.
    \item[] Justification: We describe the model architecture, routing variants, training setup, datasets, and evaluation benchmarks in the main paper and Appendix. We will release code and instructions to reproduce the main experimental results.
    \item[] Guidelines:
    \begin{itemize}
        \item The answer \answerNA{} means that the paper does not include experiments.
        \item If the paper includes experiments, a \answerNo{} answer to this question will not be perceived well by the reviewers.
        \item If the contribution is a dataset and\slash or model, the authors should describe the steps taken to make their results reproducible or verifiable. 
        \item Reproducibility can be provided via code, data, instructions, model checkpoints, hosted models, or other appropriate means.
        \item NeurIPS requires submissions to provide some reasonable avenue for reproducibility.
    \end{itemize}

\item {\bf Open access to data and code}
    \item[] Question: Does the paper provide open access to the data and code, with sufficient instructions to faithfully reproduce the main experimental results, as described in supplemental material?
    \item[] Answer: \answerYes{}.
    \item[] Justification: All datasets used in the paper are publicly available, and we will release the codebase with instructions for training and evaluation. At submission time, any released materials will be anonymized as required.
    \item[] Guidelines:
    \begin{itemize}
        \item The answer \answerNA{} means that paper does not include experiments requiring code.
        \item Please see the NeurIPS code and data submission guidelines for more details.
        \item While code and data release is encouraged, \answerNo{} is acceptable with justification.
        \item The instructions should contain the exact command and environment needed to reproduce the results.
        \item The authors should provide instructions on data access and preparation.
        \item The authors should provide scripts to reproduce all experimental results.
        \item At submission time, released materials should be anonymized if applicable.
    \end{itemize}

\item {\bf Experimental setting/details}
    \item[] Question: Does the paper specify all the training and test details necessary to understand the results?
    \item[] Answer: \answerYes{}.
    \item[] Justification: The main paper describes the backbone models, vision encoders, datasets, benchmarks, and routing variants. Additional training details, hyperparameters, and evaluation settings are provided in the Appendix.
    \item[] Guidelines:
    \begin{itemize}
        \item The answer \answerNA{} means that the paper does not include experiments.
        \item The experimental setting should be presented in the core of the paper to a level of detail necessary to appreciate the results.
        \item The full details can be provided either with the code, in appendix, or as supplemental material.
    \end{itemize}

\item {\bf Experiment statistical significance}
    \item[] Question: Does the paper report error bars suitably and correctly defined or other appropriate information about the statistical significance of the experiments?
    \item[] Answer: \answerNo{}.
    \item[] Justification: We do not report statistical significance tests or error bars. The experiments are computationally expensive, and our comparisons follow standard benchmark evaluation protocols for MLLMs.
    \item[] Guidelines:
    \begin{itemize}
        \item The answer \answerNA{} means that the paper does not include experiments.
        \item The authors should answer \answerYes{} if the results are accompanied by error bars, confidence intervals, or statistical significance tests.
        \item The factors of variability captured by the error bars should be clearly stated.
        \item The method for calculating the error bars should be explained.
        \item The assumptions made should be given.
        \item It should be clear whether the error bar is standard deviation or standard error.
        \item For asymmetric distributions, authors should avoid inappropriate symmetric error bars.
        \item If error bars are reported, the text should explain how they were calculated.
    \end{itemize}

\item {\bf Experiments compute resources}
    \item[] Question: For each experiment, does the paper provide sufficient information on the computer resources needed to reproduce the experiments?
    \item[] Answer: \answerYes{}.
    \item[] Justification: We provide compute resource details, including GPU type, memory, and training/evaluation cost, in the Appendix.
    \item[] Guidelines:
    \begin{itemize}
        \item The answer \answerNA{} means that the paper does not include experiments.
        \item The paper should indicate the type of compute workers, including relevant memory and storage.
        \item The paper should provide the amount of compute required for individual runs and estimate total compute.
        \item The paper should disclose whether the full research project required more compute than the experiments reported.
    \end{itemize}

\item {\bf Code of ethics}
    \item[] Question: Does the research conducted in the paper conform, in every respect, with the NeurIPS Code of Ethics?
    \item[] Answer: \answerYes{}.
    \item[] Justification: The research uses publicly available datasets and models, does not involve human subjects, and does not require deviations from the NeurIPS Code of Ethics.
    \item[] Guidelines:
    \begin{itemize}
        \item The answer \answerNA{} means that the authors have not reviewed the NeurIPS Code of Ethics.
        \item If the authors answer \answerNo{}, they should explain the special circumstances that require a deviation.
        \item The authors should make sure to preserve anonymity.
    \end{itemize}

\item {\bf Broader impacts}
    \item[] Question: Does the paper discuss both potential positive societal impacts and negative societal impacts of the work performed?
    \item[] Answer: \answerNA{}.
    \item[] Justification: This work is foundational research on visual representation routing in MLLMs and is not tied to a specific deployment or application domain. We do not identify a direct societal impact beyond the general impacts associated with improving multimodal model capabilities.
    \item[] Guidelines:
    \begin{itemize}
        \item The answer \answerNA{} means that there is no societal impact of the work performed.
        \item If the authors answer \answerNA{} or \answerNo{}, they should explain why.
        \item Examples of negative societal impacts include malicious uses, fairness considerations, privacy considerations, and security considerations.
        \item Foundational research may not need application-specific broader impact discussion unless there is a direct path to harm.
        \item Authors should consider harms from intended use, incorrect results, and misuse.
        \item If there are negative societal impacts, authors may discuss mitigation strategies.
    \end{itemize}

\item {\bf Safeguards}
    \item[] Question: Does the paper describe safeguards that have been put in place for responsible release of data or models that have a high risk for misuse?
    \item[] Answer: \answerNA{}.
    \item[] Justification: The paper does not release a new high-risk dataset or pretrained generative model. The released code is intended to reproduce the proposed routing method and experiments using existing public assets.
    \item[] Guidelines:
    \begin{itemize}
        \item The answer \answerNA{} means that the paper poses no such risks.
        \item Released models with high misuse risk should be released with necessary safeguards.
        \item Datasets scraped from the Internet could pose safety risks.
        \item Authors are encouraged to make a best faith effort.
    \end{itemize}

\item {\bf Licenses for existing assets}
    \item[] Question: Are the creators or original owners of assets used in the paper properly credited and are the license and terms of use explicitly mentioned and properly respected?
    \item[] Answer: \answerYes{}.
    \item[] Justification: We cite the original papers for all datasets, models, and codebases used in this work. We use publicly available assets and will document their licenses and terms of use in the Appendix or released codebase.
    \item[] Guidelines:
    \begin{itemize}
        \item The answer \answerNA{} means that the paper does not use existing assets.
        \item The authors should cite the original paper that produced the code package or dataset.
        \item The authors should state which version of the asset is used and, if possible, include a URL.
        \item The name of the license should be included for each asset.
        \item For scraped data, copyright and terms of service should be provided.
        \item If assets are released, license, copyright information, and terms of use should be provided.
        \item For repackaged datasets, both original and derived licenses should be provided.
        \item If license information is unavailable, authors are encouraged to contact asset creators.
    \end{itemize}

\item {\bf New assets}
    \item[] Question: Are new assets introduced in the paper well documented and is the documentation provided alongside the assets?
    \item[] Answer: \answerYes{}.
    \item[] Justification: We introduce and release the \molfull{} codebase. The released asset will include documentation, setup instructions, training and evaluation scripts, and license information.
    \item[] Guidelines:
    \begin{itemize}
        \item The answer \answerNA{} means that the paper does not release new assets.
        \item Researchers should communicate details of the dataset\slash code\slash model through documentation.
        \item The paper should discuss whether and how consent was obtained from people whose asset is used.
        \item At submission time, assets should be anonymized if applicable.
    \end{itemize}

\item {\bf Crowdsourcing and research with human subjects}
    \item[] Question: For crowdsourcing experiments and research with human subjects, does the paper include the full text of instructions given to participants and screenshots, if applicable, as well as details about compensation?
    \item[] Answer: \answerNA{}.
    \item[] Justification: This paper does not involve crowdsourcing or research with human subjects.
    \item[] Guidelines:
    \begin{itemize}
        \item The answer \answerNA{} means that the paper does not involve crowdsourcing nor research with human subjects.
        \item Including this information in supplemental material is fine.
        \item According to the NeurIPS Code of Ethics, workers involved in data collection, curation, or other labor should be paid at least minimum wage.
    \end{itemize}

\item {\bf Institutional review board (IRB) approvals or equivalent for research with human subjects}
    \item[] Question: Does the paper describe potential risks incurred by study participants, whether such risks were disclosed to the subjects, and whether IRB approvals were obtained?
    \item[] Answer: \answerNA{}.
    \item[] Justification: This paper does not involve crowdsourcing or research with human subjects, so IRB approval is not applicable.
    \item[] Guidelines:
    \begin{itemize}
        \item The answer \answerNA{} means that the paper does not involve crowdsourcing nor research with human subjects.
        \item IRB approval or equivalent may be required depending on the country or institution.
        \item Authors should adhere to the NeurIPS Code of Ethics and institutional guidelines.
        \item For initial submissions, do not include information that would break anonymity.
    \end{itemize}

\item {\bf Declaration of LLM usage}
    \item[] Question: Does the paper describe the usage of LLMs if it is an important, original, or non-standard component of the core methods in this research?
    \item[] Answer: \answerNA{}.
    \item[] Justification: The paper studies MLLM architectures, but LLMs are not used as a non-standard research assistant or data-generation component in the core methodology. Any use of LLMs for writing, editing, or formatting does not affect the scientific contribution.
    \item[] Guidelines:
    \begin{itemize}
        \item The answer \answerNA{} means that the core method development in this research does not involve LLMs as any important, original, or non-standard components.
        \item Please refer to the LLM policy in the NeurIPS handbook for what should or should not be described.
    \end{itemize}

\end{enumerate}

\end{document}